\documentclass{article}
\usepackage{graphicx} 
\usepackage{amsmath}
\usepackage{amssymb}
\usepackage{bbm}
\usepackage{subcaption}
\usepackage[most]{tcolorbox}
\usepackage[a4paper,margin=2.5cm]{geometry}
\usepackage{float}
\usepackage{url}
\usepackage{natbib}
\usepackage{booktabs}
\usepackage{makecell}
\usepackage[normalem]{ulem}
\usepackage{dsfont}
\usepackage{colonequals}
\usepackage{tablefootnote}
\usepackage[inline]{enumitem}
\usepackage{siunitx}
\usepackage{inconsolata}

\usepackage{xcolor}
\usepackage{xspace}

\definecolor{materialgreen500}{RGB}{76,175,80}

\newtcolorbox{mybox}[2][]{%
  colframe=blue!70!black, 
  colback=blue!10!white,  
  coltitle=white,         
  title=#2,               
  #1                      
}

\newcounter{DefBoxCounter}
\newtcolorbox[use counter=DefBoxCounter]{defbox}[2][]{
  colframe=blue!70!black, 
  colback=blue!10!white,  
  coltitle=white,         
  center, 
  valign=top, 
  halign=left,
  breakable,
  enhanced,
  title=\bfseries\sffamily Definition \theDefBoxCounter~(#2) \hfill,
  #1
}

\newtcolorbox[use counter=ExampleCounter]{example}[2][]{
  colframe=black!0, 
  colback=black!10,  
  colbacktitle=black!80,
  titlerule=0pt, 
  coltitle=white,         
  title=#2,               
  breakable,
  enhanced jigsaw,
  #1                      
}
\newtcolorbox{example_code}[1][]{
  colback=black!10,  
  colframe=black!0, 
  sharp corners,
  titlerule=0pt, 
  boxrule=1pt,
  enhanced jigsaw,
  #1                      
}

\newtcolorbox[use counter=RemarkCounter]{remark}[2][]{%
  colframe=materialgreen500, 
  colback=materialgreen500!10,  
  coltitle=white,         
  title=#2,               
  #1                      
}

\newtheorem{definition}{Definition}

\makeatletter
\newcommand{\llbracket}{\mathopen{[\![}}
\newcommand{\rrbracket}{\mathclose{]\!]}}
\makeatother

\newcommand{\ive}[1]{\llbracket#1\rrbracket}
\newcommand{\qequal}[0]{\stackrel{?}{=}}
\newcommand{\defeq}{\stackrel{\text{\normalfont \sffamily def}}{=}}

\newcommand{\qq}[1]{``#1''}

\usepackage{tikz}
\usetikzlibrary{circuits.logic.IEC}
\usepackage{circuitikz}
\tikzstyle{union} = [rounded corners,text centered, draw=black]
\tikzstyle{intersection} = [rounded corners,text centered, draw=black]
\tikzstyle{times} = [circle,text centered, draw=black]
\tikzstyle{plus} = [circle,text centered, draw=black]
\tikzstyle{int} = [circle, text centered, draw=black]
\tikzstyle{leaf} = [rounded corners, text centered, draw=none, fill=none] 
\tikzstyle{arrow} = [thick,=,>=stealth]
\tikzstyle{dirarrow} = [thick,->,>=stealth]

\newcommand{\iimage}[1]{\ensuremath{\vcenter{\hbox{\includegraphics[height=11pt]{#1}}}}}

\newcommand{\digit}[1]{\iimage{figures/mnist_#1}}

\usetikzlibrary{fit, positioning}

\newcommand{\new}[1]{#1}
\newcommand{\len}[1]{}

\newcommand{\lf}[0]{\ensuremath{\alpha}}        

\newcommand{\bigboxplus}{\mathop{\scalebox{1.7}{$\boxplus$}}}

\DeclareMathOperator*{\lpif}{\mathtt{{:}{\---}}}

\DeclareMathOperator*{\prob}{\mathtt{{:}{:}}}

\newcommand{\dl}{DeepLog\xspace}

\newcommand{\nesyebms}{NeSy-EBMs\xspace}

\title{The DeepLog Neurosymbolic Machine}

\author{Vincent Derkinderen$^*$\textsuperscript{, a} \and Robin Manhaeve\footnote{First authors} \textsuperscript{, a} \and Rik Adriaensen\textsuperscript{a} \and Lucas Van Praet\textsuperscript{a} \and Lennert De Smet\textsuperscript{a} \and Giuseppe Marra$^\dagger$\textsuperscript{, a} \and Luc De Raedt\footnote{Senior authors} \textsuperscript{, a}
\\[1em]
\textsuperscript{a}%
\footnotesize{
Department of Computer Science and Leuven.AI,
KU Leuven, Belgium} \\
\footnotesize{\url{firstname.lastname@kuleuven.be}}
}

\begin{document}

\date{}
\maketitle

\begin{abstract}

We contribute  a theoretical and operational framework for neurosymbolic AI  called DeepLog. 
DeepLog introduces building blocks and primitives for neurosymbolic AI that make abstraction
of commonly used representations and computational mechanisms  in neurosymbolic AI. DeepLog can represent and emulate a wide range of neurosymbolic systems. 
It consists of two key components.
The first is the DeepLog language for specifying neurosymbolic models and inference tasks. 
This language consists of an annotated neural extension of grounded first-order logic, and 
makes abstraction of the type of logic, e.g. Boolean, fuzzy or probabilistic, and whether logic is used in the architecture or in the loss function.
The second DeepLog component is situated at the computational level and uses extended algebraic circuits as computational graphs.
Together these two components are to be considered as a neurosymbolic abstract machine, 
with the DeepLog language as the intermediate level of abstraction
and the circuits level as the computational one. 
DeepLog is implemented in software, relies on the latest insights in implementing algebraic circuits on GPUs,  
and is declarative in that it is  easy to obtain different neurosymbolic models by making different choices for the underlying algebraic structures and logics.  The generality and efficiency of the DeepLog neurosymbolic machine is demonstrated through an experimental comparison between  1) different fuzzy and probabilistic logics, 2) between using logic in the architecture or in the loss function, and 3) between a standalone  CPU-based implementation of a neurosymbolic AI system and a DeepLog GPU-based one.

\end{abstract}

\section{Introduction}
Neurosymbolic artificial intelligence is a broad research field focusing on the integration of learning and reasoning. Although there are many views on neurosymbolic artificial intelligence, we view it as combining logic-based AI techniques with neural networks and deep learning, in line with~\citet{garcez2023neurosymbolic}. 
The interest in this field has sharply risen in the past few years due to the attention for trustworthy AI~\citep{garcez2023neurosymbolic}.
Neurosymbolic AI promises to enhance the interpretability, robustness, and data-efficiency of neural network-based models by exploiting knowledge formulated in a symbolic, often logic-based manner.  

However, the field of neurosymbolic AI is fragmented: there exist numerous neurosymbolic AI systems and their commonalities and differences are poorly understood today. What is lacking is a theoretical framework for neurosymbolic AI defining what the building blocks and primitives for neurosymbolic AI are, 
and a corresponding implementation in open-source software packages that smoothly incorporates these into standard 
machine learning software. 
It is this combination of theory and software that both enabled and accelerated the deep learning revolution. 
Indeed, there exist many related open-source software packages that allow to develop neural network models~\citep{tensorflow2015-whitepaper,jax2018github,PyTorch2024}.
These packages provide composable and efficient components, thereby enabling rapid prototyping, which has led to some forms of standardization of deep learning. 
Importantly, the development of appropriate tools has gone hand-in-hand with the creation of building blocks and primitives for neural models, deepening our understanding of the core concepts underlying neural network computation.
As such, developing such software is not merely a practical task --  it also plays a crucial role in establishing robust theoretical foundations.
It is precisely this gap in the state of the art in neurosymbolic AI that we want to fill. 

\paragraph{Contribution}
More specifically, we contribute the DeepLog framework for neurosymbolic AI.
DeepLog is a neurosymbolic machine consisting of two key components:
\begin{enumerate*}[label=\textbf{(\arabic*)}]
    \item the DeepLog language that allows to express a wide range of existing neurosymbolic AI systems that combine logic and neural networks, and
    \item an underlying computational graph representation based on algebraic circuits.
\end{enumerate*}
The DeepLog language is not a high-level language, but is situated at an intermediate level that is similar to that of propositional logic in knowledge representation.
Together with the circuits, DeepLog can be viewed as a kind of abstract machine 
that can be used to construct novel and reconstruct existing neurosymbolic frameworks and systems. 
DeepLog is based on the identification and operationalization of fundamental components of neurosymbolic AI systems gathered in a series of analyses, surveys and tutorials \citep{starai_book,marra2024statistical,de2025defining} such as the logic that is used, e.g. Boolean, fuzzy, probabilistic or a combination of these, the types of query that can be answered, and the computational mechanism for performing inference. 
DeepLog can therefore be used for theoretical studies of computational aspects of neurosymbolic AI.  
To exemplify this, we illustrate how the underlying algebraic circuits can be optimized for efficiency.
DeepLog is also operational as expressions in the DeepLog language can be compiled into 
circuits that are directly usable for learning and inference.
It is implemented in software and can therefore be used
for a direct evaluation and comparison of the different choices one can make in neurosymbolic AI.
To exemplify this, we introduce an experimental comparison between different fuzzy and probabilistic logics on some well-known benchmarks.

\paragraph{Assumptions}

DeepLog is based on the following assumptions:

\begin{itemize}
\item {\bf A common three level architecture for neurosymbolic AI.} By analogy to contemporary knowledge representation and neural network languages and software, we distinguish three levels for neurosymbolic AI models and systems. 
The first is the higher-level in which the models are described, for example, Answer Set Programs in knowledge representation, and DeepProbLog~\citep{manhaeve_neural_2021} and Logic Tensor Networks~\citep{donadello2017ltn} in neurosymbolic AI.
Although they allow to define models in a compact manner, they often possess an intricate semantics that is model-specific and relies on particular language assumptions.

The second or intermediate level operationalizes the semantics of the high-level model, for instance, through propositional logic in knowledge representation,
grounded (and possibly annotated) logical theories in answer set programming, and standard layers in TensorFlow or PyTorch.   
The intermediate level encodes \textit{what} should be calculated in a more primitive and commonly used language but it does not specify \textit{how} this should be calculated.  For neurosymbolic AI systems, such an intermediate language is still missing, and DeepLog wants to fill that gap. 
The third level is the computational one. At this level, models are expressed through simple computational operations (such as conjunctions, additions, multiplications, tensor operations) typically organised in a computational graph or circuit. 

\begin{figure}[ht!]
    \centering
    \includegraphics[width=0.333\linewidth]{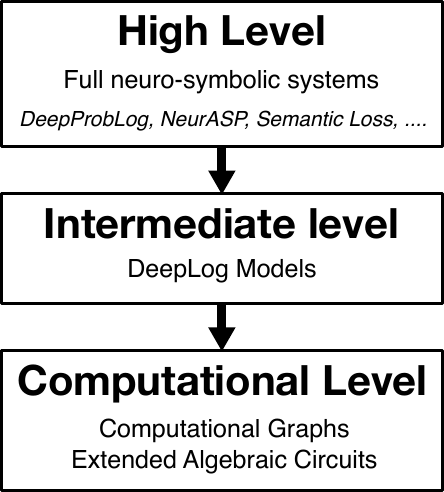}
    \caption{The three level architecture for neurosymbolic AI}
    \label{fig:levels}
\end{figure}

\item {\bf A neurosymbolic abstract machine. } 
DeepLog is not aimed at developing a high-level lingua franca for neurosymbolic AI as such an attempt is likely to fail given that there is not yet a generally agreed upon lingua franca for specifying models in knowledge representation. DeepLog rather operates at the intermediate and computational level. 
Motivated  by the prevalent use of propositional logic (or, equivalently grounded first-order logic) in knowledge representation and neurosymbolic AI, DeepLog is based on a neural extension of grounded first order logic at the intermediate level.
The computational level is based on extended algebraic circuits as computational graphs. Thus, DeepLog operates at the second and third level, and can therefore  be considered as {\em a neurosymbolic abstract machine.}

\item {\bf  Based on the unifying formal definition of neurosymbolic AI.}
DeepLog is based on the unifying formal definition of neurosymbolic AI by \citet{de2025defining}, who show that inference in numerous 
neurosymbolic AI systems can be viewed as  aggregation over the product of a logic function and a belief function. By varying the aggregation operator,
logic and belief functions, their definition makes abstraction of  many contemporary neurosymbolic AI systems, regardless of whether they are based on Boolean, fuzzy or probabilistic logics, or on a first-order or logic programming based semantics.
We adopt their theoretical definition  and turn
it into an operational and computational perspective in the DeepLog neurosymbolic machine.

\item {\bf  Labelling functions as a fundamental abstraction.}  The key step in operationalizing the definition of De Smet and De Raedt is the introduction of \textit{labelling functions}. These labelling functions, based on the model-based approach to the semantics of symbolic languages \citep{tarski1983logic, chang1990model},  assign values to interpretations of symbolic languages and provide a unified framework for various fundamental paradigms, including predicate logic, probability theory, fuzzy logic, logic programming, etc. We will demonstrate that many existing neurosymbolic frameworks and tasks can be understood as compositions of such basic labelling functions.  This also allows a natural interpretation of neural components as parametric labelling functions. For instance, DeepProbLog can be interpreted as a combination of a logic programming labelling function and a neurally-parametrized probabilistic labelling function, while Logic Tensor Networks (LTN)  correspond to a neurally-parametrized fuzzy logic labelling function. This compositional perspective within a formal framework has been largely absent in the neurosymbolic AI landscape, where most foundational work on neurosymbolic AI theory has tended to be informal and descriptive \citep{marra2024statistical, box}, or limited to specific classes of neurosymbolic systems \citep{dickens2024mathematical}.

Moreover, this compositional perspective draws a parallel between labelling functions in neurosymbolic systems and {Layer}s (or {Module}s) in deep learning frameworks, like PyTorch or Tensorflow.  By formalizing labelling functions as an abstraction, DeepLog enables the modular design of neurosymbolic frameworks. This abstraction allows for the reuse and recombination of existing labelling functions while preserving the flexibility to define novel ones.

\end{itemize}
Along with the definition of the DeepLog neurosymbolic machine at the intermediate and computational level, we contribute the \emph{DeepLog} software\footnote{The DeepLog software will be made publicly available upon acceptance of the paper.}: a   package that implements the DeepLog theory targetting neurosymbolic AI. DeepLog's flexible and efficient components result in a common  and powerful platform for implementing novel and existing neurosymbolic AI systems. This in turn facilitates empirical investigations into different components of neurosymbolic AI systems.

\paragraph{Summary}
We contribute \emph{DeepLog}, an abstract neurosymbolic machine that functions both as a theoretical framework and as a practical software package. 
 It operates at the intermediate and computational level of neurosymbolic AI.

More specifically, the distinguishing features of DeepLog are that
\begin{enumerate}
\item  DeepLog consists of {\em a formal language for defining neurosymbolic systems}; the DeepLog language is based on a  neurally extended grounded first-order logic;
\item  DeepLog offers {\em a computational framework for inference and learning}, based on extended algebraic circuits as computational graphs;
\item  DeepLog models are {\em mathematically defined so that they can be computationally analysed and used for optimisation}, 
by reasoning about and restructuring its circuits and computational graph; 
\item  DeepLog is {\em operational and implemented in software};
\item  DeepLog can {\em emulate a wide range of neurosymbolic AI systems}, as we shall show in the experimental section;
\item  DeepLog is {\em modular} as DeepLog models and components can be composed
in a flexible manner. 
\end{enumerate}

As DeepLog allows to reconstruct and emulate many existing neurosymbolic systems, it provides a novel unifying perspective on the field, 
which should pave the way towards an operational understanding of the commonalities and differences of neurosymbolic systems, and consequently, towards a deeper understanding of the key concepts and primitives underlying neurosymbolic AI.

\section{Illustrating the DeepLog Framework}\label{sec:illustration}

In this section, we  provide a first illustration of the three levels of neurosymbolic AI, focusing on an intuitive explanation of the DeepLog neurosymbolic machine. Some details of the model will become clearer only later in this work.

As a running example, we will consider variations
of the task of predicting whether an alarm should be activated. We know that the alarm must be activated when there is a \texttt{burglary} or an \texttt{earthquake}. In addition to this knowledge, there is data: the videostream from a camera (denoted as variable \texttt{Video}), and time series data from a seismic sensor (denoted as variable \texttt{Seismic}). 
The approach is to learn a neural detection model for each concept, i.e. for \texttt{burglary(Video)}, for \texttt{earthquake(Seismic)}, and possibly for some other predicates  that we may be using later on (such as \texttt{alarm(Video,Seismic)}). 

A neurosymbolic  model  could be used (and trained) to query for the probability that a specific  
video instance \iimage{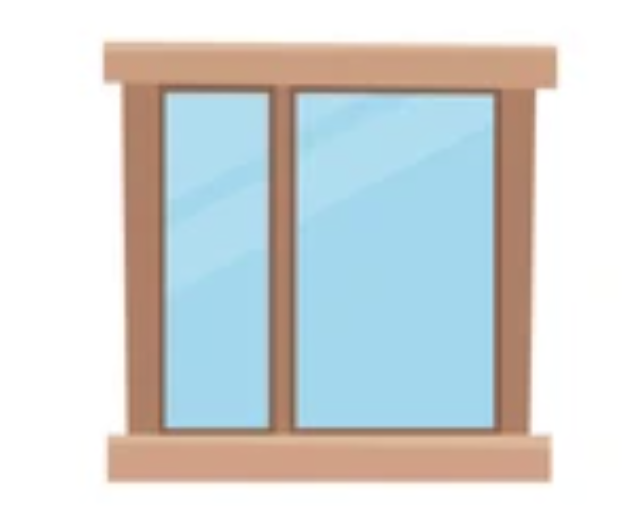} and seismic data \iimage{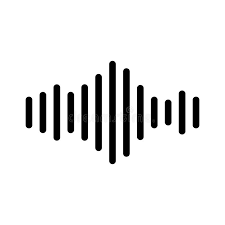} indicate an alarming situation, which occurs when \(\mathtt{burglary(\iimage{images/burglary1})} \vee \mathtt{earthquake(\iimage{images/earthquake1})}\) holds. 
A high-level modeling language that could be used to represent this is DeepProblog~\citep{manhaeve2018deepproblog}. The model would look as follows. For more info on the syntax, we refer to Appendix~\ref{app:syntax}. 
\begin{example_code}


\noindent
\texttt{nn(classifier1, Video) :: burglary(Video).} \\
\texttt{nn(classifier2, Seismic) :: earthquake(Seismic).} \\
\texttt{?- burglary($\iimage{images/burglary1}$) ; earthquake($\iimage{images/earthquake1}$).}
\end{example_code}

The purpose of this example is to illustrate how the neural networks {\tt classifier1} and {\tt classifier2} can be used to estimate the probabilities of a {\tt burglary} and an {\tt earthquake}, respectively.  
Let $e$ represent earthquake and $b$ represent burglary. We can define the probability that there is a burglary or an earthquake using the following components: 
\begin{itemize}
\item
Defining the logic formula: 
\(\varphi = \texttt{b}(V) \vee \texttt{e}(S)\)
where the idea is that $\texttt{b}(V)$ and $\texttt{e}(S)$ are Boolean atoms that can be assigned values
 $true$ and $false$ in a possible world once a video $V$ and a seismic sensor reading $S$ have been provided. 
 \new{To explicitly aggregate and refer to those truth values, we introduce two reified variables to represent them, $B$ and $E$ respectively. This results in our logic DeepLog formula} \[\varphi(B,E,V,S) = \texttt{b}(V)[B] \lor \texttt{e}(S)[E].\]
\item
Defining the belief function:
\(Prob(B,E,V,S) = Prob(B,V) \times Prob(E,S)\), with 
\begin{align*}
    Prob(B,V) &= 
    \begin{cases}
        nn_{1}(V) &\text{ if $B$ is true,} \\
        1-nn_{1}(V) &\text{ otherwise}
    \end{cases} \\
    Prob(E,S) &=
    \begin{cases}
        nn_{2}(S) &\text{ if $E$ is true,}\\
        1-nn_{2}(S) &\text{ otherwise}
    \end{cases}
\end{align*}
where $nn_{i}$ refers to the neural networks that predict the probabilities of $\texttt{b}(V)$ and $\texttt{e}(S)$ being true, using $V$ and $S$ as inputs; 
\item 
The probability of the formula $\varphi$ holding in any possible world is then the aggregation over all the possible assignments of $true$ and $false$:
  \begin{equation}
    \label{eq-il}
          \sum_{B \in \{t,f\}} \sum_{E \in \{t,f\}} \varphi(B,E,V,S) \times Prob(B,E,V,S)
    \end{equation}
\end{itemize}

While the example above relies on Boolean logic and a probabilistic belief function, alternative choices are possible, for instance using fuzzy logic or a hybrid fuzzy-probabilistic approach, cf. \cite{de2025defining,marra2024statistical} and below. These choices are made explicit in the DeepLog intermediate language by specifying the algebraic structure in which they are interpreted and computed.
The core of the resulting formula in the intermediate language is as follows, where $\mathbb{B}$ and 
$\mathbb{P} $ denote the Boolean and probabilistic algebraic structures, respectively.
\begin{align*}
\varphi_\mathbb{P}(V, S) = 
    \sum_{B} \sum_{E} \texttt{b}(V)[B]_\mathbb{B} \lor_\mathbb{B}
        \texttt{e}(S)[E]_\mathbb{B})_\mathbb{P} \times_\mathbb{P}
        (\texttt{b}(V)[B]_\mathbb{P} \times_\mathbb{P} \texttt{e}(S)[E]_\mathbb{P})
\end{align*}
\new{This defines the probability that $\texttt{b}(V)$ or $\texttt{e}(S)$ is true, as a function $\varphi$ of input $V$ and $S$.}
The advantage of this explicit formulation is that it facilitates the modeling of alternative semantics and reasoning systems by simply modifying the formula and the underlying algebraic structures, $\mathbb{B}$ and $\mathbb{P}$. For instance, one can transition from DeepProbLog to a probabilistic fuzzy logic semantics \citep{marra2019integrating, NeuPSLPryor23} by replacing the Boolean logic structure $\mathbb{B}$ with a t-norm-based fuzzy logic structure $\mathbb{F}$, while keeping the rest of the formalization and computational procedures unchanged. This substitution effectively captures the \textit{unique} core difference between the frameworks, while preserving all other components. We view this modularity as a fundamental feature when aiming to advance the development of neurosymbolic AI.

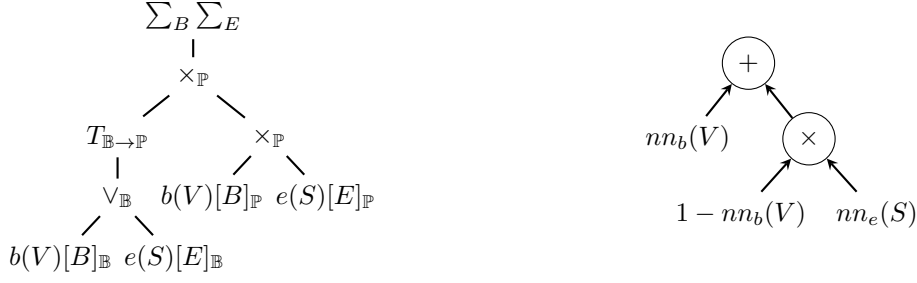
\begin{figure}
    \centering
    \begin{minipage}{.48\linewidth}
        \centering
        \begin{tikzpicture} [node distance=0.8cm]
    \node (n1) [leaf] {$\sum_{B} \sum_{E}$};
    \node (n2) [leaf, below of=n1] {$\times_\mathbb{P}$};
    \node (n3) [leaf, below of=n2, right of=n2, xshift=0.2cm] {$\times_\mathbb{P}$};
    \node (n4) [leaf, below of=n2, left of=n2, xshift=-0.2cm] {$T_{\mathbb{B} \rightarrow \mathbb{P}}$};
    \node (n5) [leaf, below of=n3, left of=n3, xshift=0.05cm] {$b(V)[B]_\mathbb{P}$};
    \node (n6) [leaf, below of=n3, right of=n3, xshift=-0.05cm] {$e(S)[E]_\mathbb{P}$};
    \node (n7) [leaf, below of=n4] {$\lor_\mathbb{B}$};
    \node (n8) [leaf, below of=n7, left of=n7, xshift=0.05cm] {$b(V)[B]_\mathbb{B}$};
    \node (n9) [leaf, below of=n7, right of=n7, xshift=-0.05cm] {$e(S)[E]_\mathbb{B}$};
    
    \draw [arrow] (n2) -- (n1);
    \draw [arrow] (n3) -- (n2);
    \draw [arrow] (n4) -- (n2);
    \draw [arrow] (n5) -- (n3);
    \draw [arrow] (n6) -- (n3);
    \draw [arrow] (n7) -- (n4);
    \draw [arrow] (n8) -- (n7);
    \draw [arrow] (n9) -- (n7);
\end{tikzpicture}
  
    \end{minipage}
    \begin{minipage}{.48\linewidth}
        \centering
         \begin{tikzpicture} [node distance=1cm]

    \node (n1) [plus] {$+$};
    \node (na) [leaf, below of=n1, left of=n1, xshift=0.2cm] {$nn_b(V)$};
    \node (n2) [times, below of=n1, right of=n1, xshift=-0.2cm] {$\times$};

    \node (nna) [leaf, below of=n2, left of=n2, xshift=0.1cm] {$1 - nn_b(V)$};
    \node (nb) [leaf, below of=n2, right of=n2, xshift=-0.1cm] {$nn_e(S)$};

    \draw [dirarrow] (na) -- (n1);
    \draw [dirarrow] (n2) -- (n1);
    \draw [dirarrow] (nna) -- (n2);
    \draw [dirarrow] (nb) -- (n2);
\end{tikzpicture}
  
    \end{minipage}
    \caption{Formula (left) and the corresponding optimized computational circuit (right) for the running example, predicting whether a burglary or earthquake occurs.}
    \label{fig:intro-circuit-exa}
\end{figure}

The DeepLog intermediate level can be mapped onto the computational one  
using algebraic circuits as computational graphs,
which make the elementary operations and computations explicit. They can be used for inference and learning through differentiation. The computational graph that corresponds to our running example, and the optimized circuit for it, are both shown in Figure~\ref{fig:intro-circuit-exa} and will be explained in Section~\ref{sec:composing_ac}.

\section{Analysing Neurosymbolic AI}\label{sec:analysation}

The field of neurosymbolic AI is broad and encompasses a wide range of approaches.  Although there is not yet a universally accepted
theory of neurosymbolic AI, \citet{marra2024statistical} analyse  state-of-the-art models and techniques along a number of important dimensions
that we now summarize as an introduction to the field. 
Throughout this paper, we assume basic familiarity with first-order logic and logic programming. For an in-depth explanation, we refer to \citet{flach:simplylogical}.

\subsection{Constraints vs. Rules}
One of the most fundamental divisions in neurosymbolic AI arises from differing interpretations of logic — whether logic sentences should be viewed as a system of constraints or as a set of rules or definitions.

\paragraph{Constraints.} In the first perspective, logic formulas are treated as constraints over a set of variables, and reasoning is framed as the process of finding variable assignments that satisfy these constraints, much like in satisfiability (SAT) problems. This boils down to a model-theoretic perspective on logic. A large class of neurosymbolic systems adopts this approach by leveraging neural networks to output values for the variables for which satisfying assignments are sought. In these models, logic constraints are incorporated \textbf{as loss functions}, guiding the training of neural networks. As a result, reasoning is effectively reinterpreted as a learning process: rather than explicitly searching for valid assignments, the model is trained to penalize those assignments that violate the given constraints. This approach embeds knowledge as a form of regularization, complementing standard supervised learning by softly enforcing logical consistency.

\begin{example}{\sffamily Constraints}
Consider the task above of predicting whether an \texttt{alarm} should be activated. 
In the constraint approach, one could learn a model for each individual concept: one for \texttt{alarm(Video,Seismic)}, one for \texttt{burglary(Video)} and one for \texttt{earthquake(Seismic)}. We can enhance those models by using the additional knowledge---the conditions under which an alarm should be activated. In practice, this means incorporating the logical constraint below into a differentiable loss function that penalises inconsistent predictions, i.e. those predictions that are inconsistent with respect to the constraint.
\[
    \forall~\texttt{Video}, \texttt{Seismic}: \;
    \texttt{alarm(Video,Seismic)} \leftrightarrow 
    (\texttt{burglary(Video)} \lor \texttt{earthquake(Seismic)})
\]
\end{example}

\paragraph{Rules as definitions.}
The second perspective views logical rules as definitions, i.e. the rules define predicates in terms of other predicates, and the logic becomes a kind of formal rewrite system. In this perspective, reasoning means iteratively applying these rules to derive conclusions (the defined symbol) from given premises (the symbols used in the definition). This perspective more naturally adheres to the proof-theoretic perspective in logic.  Another major class of neurosymbolic AI systems follows this paradigm, using neural networks to compute premises that are then processed through logical definitions to infer conclusions. Differentiable instantiations of these inference steps enable logic to be integrated directly into the \textbf{forward pass of the neural architecture}, effectively functioning as an additional computational layer \citep{de2020statistical}.

\begin{example}{\sffamily Definitions}
Consider now the case in which \texttt{alarm(Video,Seismic)} is not explicitly modelled but we can use the additional knowledge to infer its value from the other predicted concepts. This is a hybrid model where we first have neural models for \texttt{burglary(Video)} and \texttt{earthquake(Seismic)}, which are then combined explicitly using an additional neurosymbolic layer to predict \texttt{alarm(Video,Seismic)}. This neurosymbolic layer is based on the same logical rule, but seen as a definition of the left-hand side atom \texttt{alarm(Video,Seismic)}, now written as a logic program: 
\begin{example_code}
[colframe=black!10]
\noindent 
\texttt{alarm(Video,Seismic) :- burglary(Video).} 

\texttt{alarm(Video,Seismic) :- earthquake(Seismic).}
\end{example_code}

In other words, instead of predicting {\tt alarm} independently and then check its consistency w.r.t. {\tt burglary} and {\tt earthquake}, we directly compute {\tt alarm} based on them. Therefore, this constraint is enforced by construction of the architecture and the predictions are guaranteed to adhere to the rule. This is not the case for the constraint approach.
\end{example}

In both cases, logic—and the knowledge it represents—serves as an inductive bias in learning. When logic is treated as constraints, this bias is soft, as the constraints are relaxed into differentiable penalties. This relaxation allows for computational efficiency during inference, enabling the use of complex logical constraints during training while maintaining tractability for inference.  
Conversely, when logic is incorporated as rules or definitions, it imposes a hard bias, directly structuring the reasoning process within the model as an additional computation.

\subsection{Syntax and Semantics}
Within knowledge representation, there exists a wide range of logics that are used in artificial intelligence. They typically build upon first-order logic and can be distinguished 
along the following dimensions.

\paragraph{Propositionalization.} In a first-order logic, one can use variables and quantifiers in expressions. In a propositional logic the predicates do not have arguments, that is, they are of arity 0. Throughout this paper, we will consider ground atoms (i.e. atoms that do not contain variables) as propositions. Nearly all neurosymbolic models define their logical semantics—including probabilistic and fuzzy extensions—through propositional representations. Consequently, effective techniques are required to translate first-order logical theories and logic programs into equivalent propositional theories. This translation is usually performed through \textit{grounding}. Grounding a first-order theory is the process of systematically substituting all variables by all possible constants (or ground terms). Grounding a first-order logic formula effectively reduces it to a set of propositional formulae, one for each substitution. This process can generate an exponential number of  propositional formulae, posing significant computational challenges. An alternative to grounding when performing inference, is to  employ \textit{theorem-proving} techniques that selectively instantiate only the subset of propositions relevant to a given query. This approach significantly reduces the computational burden, making reasoning more efficient in practice.

\begin{example}{\sffamily Grounding vs theorem-proving}
One well-known example in the neurosymbolic community is that of the MNIST addition, where the classification of two individual digits are summed. For this example, we can consider the two images to be constant tensors \texttt{i1} and \texttt{i2}.
For the variables, we have the two classifications \(\mathtt{D1},\mathtt{D2}\in\{0,\ldots,9\}\) and all possible sums \(\mathtt{S}\in\{0,\ldots,18\}\).

We have the following rule:
\begin{align*}
    \forall~\texttt{D1},\texttt{D2},\texttt{S}~:~&\texttt{addition(i1,i2,S)} \leftrightarrow\\
    &\texttt{digit(i1,D1)} \land \texttt{digit(i2,D2)} \land \texttt{sums(D1,D2,S)}
\end{align*}
and the following facts: \(\texttt{sums(0,0,0), sums(1,0,1),\ldots,sums(9,9,18)}\).
When grounding out the rule, we get \(10\times10\times19=1900\) instantiations.
When using theorem proving, we only get the relevant instantiations: 1 instantiation for \texttt{addition(i1,i2,0)}, 5 for \texttt{addition(i1,i2,4)}, etc. 
\end{example}

\paragraph{Semantics.} Once a propositional theory has been obtained, neurosymbolic systems further diverge in how they interpret this representation, specifically in their underlying semantics or meaning.
In systems based on full first-order logic, semantics is typically defined in terms of all possible models—that is, all assignments of truth values to logical variables that satisfy the given theory. This interpretation aligns with the open-world assumption, where facts that cannot be proven are considered unknown rather than false. 
In contrast, logic programming approaches adopt a more restrictive semantic framework, typically defined in terms of the unique minimal model (for  definite clausal logic). This model represents the assignment that satisfies the theory while minimizing the number of propositions assigned as true. This approach is closely associated with the closed-world assumption, where facts that cannot be proven are assumed to be false. Alternatives semantics, e.g. the \textit{stable model semantics} of Answer Set Programming (ASP), take  different approaches for more expressive fragments of logic programming. While these differences are very well known in the logic and logic programming literature, their subtle differences have been less exploited in the neurosymbolic frameworks. Similar logical expressions may target different models in  different semantics and, therefore, they can result in very different learning problems. In other words, two syntactically identical logical formalisms (and formulae) that are interpreted in a different semantics may lead to different neurosymbolic systems based on very different neural networks and computational graphs.

\subsection{Fuzzy vs Probabilistic}
A last key differentiating factor in the context of neurosymbolic AI is the extension of classical discrete (Boolean) semantics to continuous domains. This extension is crucial for integrating the reasoning capabilities of logic with the learning mechanisms of neural networks. We can categorize systems into two broad classes: fuzzy logics and probabilistic logics.

\paragraph{Fuzzy logic.} Fuzzy logic systems generalize Boolean semantics by allowing any value within the interval  $[0,1]$, rather than being restricted to True or False. This framework is grounded in t-norm theory, where logical conjunction is interpreted as a t-norm function  $t:[0,1] \times [0,1] \to [0,1]$. Fuzzy logic introduces an \textit{alternative} logical semantics distinct from Boolean logic. This framework has different properties, satisfies different theorems, and was originally developed to handle vagueness. Many properties that hold in Boolean logic do not necessarily hold in fuzzy logic. While this distinction is significant, the application of fuzzy logic in neurosymbolic AI is often relatively naive: it is typically regarded as a continuous relaxation of existing Boolean logic theories. This has the advantage of enabling differentiability in a computationally efficient way, but comes at the cost of a less transparent semantic interpretation and weaker connections to the original Boolean theory, certainly when multistep inference and chaining is  performed.

\paragraph{Probabilistic logic.} Probabilistic logic systems, in contrast, take a different approach by defining probability distributions over discrete structures. These systems can be further divided into two major paradigms: those that assign probability distributions over models (model-theoretic perspective) and those that assign probability distributions over proofs or logical derivations (proof-theoretic perspective). Different from fuzzy logic, probabilistic logic functions as an \textit{additional} semantic layer built upon Boolean logic, rather than as an alternative to Boolean logic.  
Unlike fuzzy logic, which addresses vagueness, probabilistic logic is designed to handle uncertainty, which aligns more naturally with the machine  learning principles in neurosymbolic AI.

\section{Unifying Neurosymbolic AI}\label{sec:unification}
In a recent paper~\citep{de2025defining}, neurosymbolic AI models are defined formally in a way that abstracts the inference tasks and representations of well-known neurosymbolic AI systems, such as SPL~\citep{ahmed2022semantic}, NeurASP~\citep{yang2020neurasp}, LTN~\citep{donadello2017ltn}, and others. 
In the present paper, we adopt and operationalize this mathematical framework.
We will now first present the intuitions behind their framework, and then, in Section~\ref{sec:generalization}, we discuss our generalisation.

A key observation that inspired their mathematical framework is that ``the vast majority of neurosymbolic AI models combine \emph{logic} with \emph{beliefs}''.
This observation naturally led to the view of neurosymbolic inference as aggregation $\int$ of the product of a
logic function $l$ and a belief function $b_{\boldsymbol{\theta}}$ over logical interpretations $I$, which
they formalize as:

\begin{equation}\label{eq:nesy_functional}
    F_{\boldsymbol{\theta}}(\varphi)
    =
    \int_{I \in \Omega}
    l(\varphi, I)
    \
    b_{\boldsymbol{\theta}}(\varphi, I)
    \ \mathrm{d} I.
\end{equation}
For the purposes of the present paper, it is not necessary to understand the full technical details of the notation and the mathematical definition as 
the DeepLog language employs a more operational notation for neurosymbolic inference. 
So, for ease of exposition, we will only provide an intuitive explanation
and illustration of their framework and refer to 
\citet{de2025defining} for technical details.

\begin{itemize}
\item 
The logic function $l$ operates over sentences \(\varphi\) in any logical language $L$, including very expressive logics such as first-order logic and logic programs, and logical interpretations (possible worlds) $I$ for the formula $\varphi$. $l(\varphi,I)$ then denotes the (fuzzy or boolean)  truth-value of the formula $\varphi$ in the interpretation $I$. 
\item 
The belief  $b_{\boldsymbol{\theta}}(\varphi,I)$ can be interpreted as a weight indicating the degree of belief (or probability) that the logic sentence $\varphi $ is satisfied under the given interpretation $I$. The belief function $b_{\boldsymbol{\theta}}$ is parametrized by the weights \(\boldsymbol{\theta}
\) of a neural network (or probabilistic model). 

\item The aggregation operator $\int_{I\in \Omega}$ typically computes a kind of expectation of the product. Thus the function $F_{\theta}(\varphi)$
can be interpreted as the expected belief that the formula is satisfied in a possible world from $\Omega$. 
\end{itemize}
The reader familiar with statistical relational artificial intelligence, or neural probabilistic logics
such as Semantic Loss~\citep{xu2018semantic} and DeepProbLog~\citep{manhaeve_neural_2021}, may 
recognize the resemblance to weighted model counting and weighted model integration \citep{ChaviraD08,wmi,KimmigBR17AMC}. 

At this point it is instructive to revisit the example provided in Section 2,
where the aggregation, logic and belief functions are made explicit:
\begin{equation*}
          \sum_{B \in \{t,f\}} \sum_{E \in \{t,f\}} \varphi(B,E,V,S) \times Prob(B,E,V,S).
\end{equation*}
This equation is an aggregation over all possible interpretations over the atoms
$\texttt{b}(V)$ and $\texttt{e}(S)$\new{, which is achieved by introducing and aggregating over reification variables $B$ and $E$,} while the formula $\varphi(B,E,V,S)$ denotes the disjunction of the two atoms.
So the logic part holds for the interpretations $\{B \mapsto t, E \mapsto t\}$,  $\{B \mapsto t, E \mapsto f\}$ and $\{B \mapsto f, E \mapsto t\}$. 
The belief function $Prob(B,E,V,S)$ is a combination of two neural networks, $nn_{1}(V)$ and $nn_{2}(S)$.

\section{Generalizing Neurosymbolic AI}\label{sec:generalization}

In Section~\ref{sec:unification}, we discussed the definition of neurosymbolic inference as a means to semantically unify a wide range of neurosymbolic AI methods.
While this definition provides a powerful abstraction, we identify several aspects that motivate a further operational generalization.

First, the belief function \(b_{\boldsymbol{\theta}}\) and logic function \(l\) share the same signature, suggesting that they are not fundamentally different but rather instances of a common underlying building block: labelling the value of a formula. We formalise this idea through the notion of a \textit{labelling function}.

Second, the restriction to exactly two functions combined by a single aggregation operation appears unnecessarily strict. Some settings may require reasoning about multiple interacting concepts, which need to be valued using nested aggregations. To address this, we introduce a language that supports arbitrary combinations and compositions of labelling functions.

Third, the abstract definition treats belief and logic functions as black boxes, ignoring their internal structure. This abstraction is not operational: it requires integration over the entire space of interpretations, which is generally infeasible in practice. Prior work in statistical relational AI~\citep{starai_book} demonstrates that exploiting the internal logical structure of such functions, as well as the properties of their composing operators, is essential for tractable reasoning and learning. Our proposed generalization makes these internal structures more explicit and composable.

\new{This generalisation is enabled by DeepLog as a framework that algebraically composes core logical concepts (Table~\ref{tab:language_examples}) into computational tasks.
These logical concepts can be labelled in different algebraic structures, e.g. probability, fuzzy or Boolean, to allow DeepLog to emulate and unify the alphabet soup of neurosymbolic systems.
Each one is outlined in detail in the following sections.
}

\begin{table}[ht!]
    \centering
    \caption{Notation of core concepts in DeepLog.}
    \label{tab:language_examples}
    \begin{tabular}{p{8cm}@{} @{}l}
        \toprule
        \textbf{Concept}
        &   \textbf{Example}
        \\        
        \toprule
        algebraic structure
        &   \(\mathbb{B} = (\{true, false\}, \{ \neg \}, \{\lor, \land\})\)
        \\
        \new{constant}
        &   \(green\)
        \\
        variable
        &   \(Video\)
        \\
        variable assignment
        &   \(\{Video \mapsto t_1, Color \mapsto green, Img \mapsto img1 \dots\}\)
        \\
        \new{reification variable}
        &   \(B\)
        \\
        \midrule
        algebraic atom
        &

        \\
        \hspace{2em} - without arguments
        &   \( \texttt{burglary}_R \)
        \\
        \hspace{2em} - with tensor constant
        &   \( \texttt{burglary}(\iimage{images/burglary1})_R \)
        \\
        \hspace{2em} - with variable
        &   \(\texttt{burglary}(Video)_R\)
        \\
        \hspace{2em} - with reification variable
        &   \( \texttt{burglary}[B]_R \)
        \\
        formula
        &
        \\
        \hspace{2em} - algebraic composition
        &   \( \texttt{burglary}_{\mathbb{B}} \lor_{\mathbb{B}} \texttt{earthquake}_{\mathbb{B}} \)
        \\
        \hspace{2em} - aggregation over a variable
        &   \(
            \sum_{Color} 
            \texttt{stoplight}(Color)_\mathbb{P} 
            \times_\mathbb{P} 
            \texttt{hasColor}(Img,Color)_\mathbb{P}
            \)
        \\
        \hspace{2em} - aggregation over a reification variable
        &   \(
            \sum_{B} \texttt{burglary}[B]_\mathbb{P}
            \times_\mathbb{P} 
            \texttt{hasColor}(Img,Color)_\mathbb{P}
            \)
        \\
        \hspace{2em} - transformation of algebraic structure
        &   \( (\texttt{burglary}_R \lor_R \texttt{earthquake}_R)_S \)
        \\
        labels
        &
        \\
        \hspace{2em} - ground algebraic atom
        &   \( \lf(\texttt{burglary}(\iimage{images/burglary1})_R) \)
        \\
        \hspace{2em} - algebraic atom with variable assignment
        &   \(\lf(\texttt{burglary}(Video)_\mathbb{P}\{Video \mapsto \iimage{images/burglary1}\})\)
        \\
        \hspace{2em} - ground formula
        &   \( \lf(\texttt{burglary}_R \lor_R \texttt{earthquake}_R) \)
        \\
        \hspace{2em} - aggregation
        &   \(
            \lf(\sum_{Color} \texttt{stoplight}(Color)_\mathbb{P}
            \times_\mathbb{P}
            \texttt{hasColor}(img1,Color)_\mathbb{P})
            \)
        \\
        \hspace{2em} - transformation
        &   \( \alpha\left( (\texttt{burglary}_\mathbb{B} \lor_\mathbb{B} \texttt{earthquake}_\mathbb{B})_\mathbb{P} \right) \)
        \\
        \bottomrule
    \end{tabular}
\end{table}

\subsection{Algebraic Extensions of Logic}\label{subsec:algebraic_structure}
Algebraic extensions of logic have played a crucial role in differentiable reasoning within neurosymbolic systems by generalizing logical inference to semirings such as the probability or the gradient semiring. 
The introduction of \emph{algebraic model counting} (AMC)~\citep{KimmigBR17AMC} has been particularly transformative, offering a unifying framework for integrating symbolic logic with numerical computation. In systems like DeepProbLog, AMC enables the seamless combination of probabilistic logic programming with gradient-based learning, facilitating end-to-end trainable models that retain symbolic structure \citep{Derkinderen24Semiring}.

\new{
AMC is not sufficient to cover all neurosymbolic systems.
Its reliance on semirings and models fits the purpose of probabilistic computations over Boolean semantics, but does not support fuzzy and some other semantic computations.
Other systems use different algebraic structures to evaluate their syntactic constructions, e.g. multivalued algebras to evaluate fuzzy logic.
}

\new{
DeepLog uses \emph{algebraic structures} to operationalise neurosymbolic AI by algebraically evaluating formulas.
The algebraic structure defines the underlying set of values (labels) of formulas and operators used to compose atoms into more complex formulas. 
Different structures then yield different evaluations corresponding to the computations of different neurosymbolic systems.
}

\begin{definition}[Algebraic Structure]
An algebraic structure \( R \) is a tuple
\[
R = (\mathcal{A}_R, \mathcal{U}_R, \mathcal{B}_R, Agg_R),
\]
where:
\begin{itemize}
    \item \( \mathcal{A}_R \) is a set of labels,
    \item \( \mathcal{U}_R \) is a set of unary operations \( u: \mathcal{A}_R \to \mathcal{A}_R \), and
    \item \( \mathcal{B}_R \) is a set of binary operations \( b: \mathcal{A}_R \times \mathcal{A}_R \to \mathcal{A}_R \).
    \item \new{\( Agg_R \) is a set of aggregation operators \(\boxplus: \mathcal{A}_R^* \to \mathcal{A}_R\).}
\end{itemize}
\end{definition}
Our examples will mainly rely on the probability semiring \(\mathbb{P} = (\mathbb{R}^+,\varnothing,\{+, \times\}, \varnothing)\), the Boolean algebra \(\mathbb{B} = ( \{true, false\}, \{ \neg \}, \{\lor, \land\}, \varnothing)\), or an algebra \(\mathbb{F}\) over fuzzy scores \(\mathcal{A}_\mathbb{F} = [0,1]\).

\new{
Algebraic structures are the necessary generalisation to support the wide range of semantics of neurosymbolic systems.
Many of these systems also have their own high-level syntax that can vary wildly.
This syntactic variation leads us to instead construct a lower-level \emph{DeepLog language} into which other languages can be translated.
}

\subsection{\new{DeepLog Formulae}}

\len{
Notational convention:
\begin{itemize}
    \item Typewriter font \texttt{$\backslash$texttt} for predicate symbols
    \item Math font for constants, variables and algebraic structures
    \item Not sure: Typewriter font \texttt{$\backslash$texttt} for truth values, e.g. \texttt{true} or fuzzy \texttt{0.7} to separate truth entities in font
\end{itemize}
}

\subsubsection{Atoms and variables}

The main building blocks of our language are \emph{algebraic atoms}.
We first present algebraic atoms in their simplest form (ground and non-reified) to ease exposition; the full definition allowing non-ground and reified variants is given shortly after.
%
%
\begin{definition}[Algebraic atom (Simplified)]
    \label{def:simple-atom}
    An algebraic ground atom is an expression of the form:
    \[
    p(t_1,\ldots,t_n)_R
    \]
    where $p$ is an n-ary predicate symbol, every $t_i$ is a constant, and $R$ is an algebraic structure.
    The latter indicates that the label associated with the algebraic atom is in \(\mathcal{A}_R\).
\end{definition}

Algebraic atoms are inductively combined with operators to form more complex logical formulae.
Logical formulae in DeepLog will be given values in algebraic structures to fulfill the goal of using various types of logics including Boolean, fuzzy and probabilistic logics.

\begin{definition}[Formula]
A DeepLog formula \( \varphi_R \) over algebraic structure \( R \) is defined inductively:
\begin{itemize}
    \item An algebraic atom \(atom_R\) is a formula \( \varphi_R \),
    \item If \( \varphi_R \) is a formula and \( u_R \in \mathcal{U}_R \), then \( u_R\, \varphi_R \) is a formula,
    \item If \( \varphi_R \) and \( \psi_R \) are formulas, and \( b_R \in \mathcal{B}_R \), then \( \varphi_R \; b_R \; \psi_R \) is a formula (we use infix notation here).
\end{itemize}
\end{definition}
Examples of algebraic atoms include \(\texttt{burglary}_\mathbb{B}\), \(\texttt{burglary}_\mathbb{P}\), and \(\texttt{hasColor}(square,green)_\mathbb{B}\).
These atoms can then be used to construct formulae such as \(\texttt{burglary}_\mathbb{P} \times_\mathbb{P} \texttt{earthquake}_\mathbb{P}\).

\new{
Neurosymbolic computation in \dl uses algebraic \emph{labels} of \dl formulae in general algebraic structures.
Just as in AMC, labels indicate the value associated with an atom.
Labels of atoms are algebraically combined into labels of formulae that indicate the value of the formula.
}

\begin{definition}[Labelling]
    \label{def:labelling}
    Let \(\varphi_R\) be a formula over an algebraic structure \(R = (\mathcal{A}_R, \mathcal{U}_R, \mathcal{B}_R)\),
    the label \(\lf(\varphi_R)\) of formula \(\varphi_R\) is defined inductively as follows.
     \begin{itemize}
     
        \item If formula \(\varphi_R\) is a ground, non-reified algebraic atom \( atom_R \), then its label is defined as
            \begin{equation}
                \lf(atom_R).
            \end{equation}
            This label is specified by the user while defining the DeepLog model.

        \item If the formula is of the form \( u_R\, \varphi_R \), then its label is defined as
            \begin{equation}
                    \lf(u_R\, \varphi_R) = u_R(\lf(\varphi_R)).
            \end{equation}

        \item If the formula is of the form \(\varphi_R \; b_R \; \psi_R  \), then its label is defined as 
            \begin{equation}
                    \lf( \varphi_R \; b_R \; \psi_R) = b_R(\lf(\varphi_R), \lf(\psi_R)).
            \end{equation}
     \end{itemize}
\end{definition}
The subscript \(R\) is important to indicate the type of label (e.g. Boolean, probability, ...), thereby distinguishing, for example, \(\lf(burglary_{\mathbb{B}})\) from \(\lf(burglary_{\mathbb{P}})\).

\begin{example}{\sffamily Simple labelling}
    Consider the formula
    \begin{equation}
        \varphi_\mathbb{P} = (\texttt{burglary}_\mathbb{P} \times_\mathbb{P} \texttt{earthquake}_\mathbb{P}),
    \end{equation}
    where \(\mathbb{P}\) refers to the probabilistic structure that contains the arithmetic multiplication \(\times\) as a binary operation over the domain of real values.
    The label of \(\varphi_\mathbb{P}\) depends on the labels of its algebraic atoms, \(\texttt{burglary}_\mathbb{P}\) and \(\texttt{earthquake}_\mathbb{P}\). As example, we consider the following labelling.
    \begin{equation*}
        \lf(\texttt{burglary}_\mathbb{P}) = 0.7; \quad \quad  \lf(\texttt{earthquake}_\mathbb{P}) = 0.99
    \end{equation*}
    Based on this labelling and the semantics of \(\times_\mathbb{P}\), it follows that the label of formula \(\varphi_\mathbb{P}\) corresponds to
    \begin{equation*}
        \lf(\varphi_\mathbb{P}) = 0.7 \times 0.99.
    \end{equation*}
\end{example}

\new{
\dl supports \emph{regular variables} to avoid excessive enumeration and allow for subsymbolic inputs.
\emph{Regular variables}, or simply variables, have a set of constants as domain much like variables in first-order logic that can represent any object of interest, from ranges of colours to images and other tensors.
Variables can appear in algebraic atoms, e.g.
$\texttt{burglary}(Video)_\mathbb{B}$ is an algebraic atom containing the variable $Video$.
An atom and a more general formula is called \emph{ground} when it contains no variables.
Atoms and formulas that are not ground can be ground by means of a \emph{variable assignment}.
}

\begin{definition}[Variable assignment]
\label{def:variable_assignment}
A variable assignment \(\sigma = \{V_1 \mapsto c_1, V_2 \mapsto c_2, \dots, V_n \mapsto c_n\}\) for a formula $\varphi_R$ is a mapping from regular variables and reification variables \(V\) in $\varphi_R$ to constants and truth values $c$ respectively.
\end{definition}


The previous definition of labelling (Definition~\ref{def:labelling}) provides the semantics of ground formulae, but it does not yet account for formulae in which variables appear. 
Such formulae are to be considered as functions; it is only after all variables have been assigned to constants that the formulae becomes ground, and that the label of the formulae is defined. 
Given a formula $\varphi_R$ and a variable assignment $\sigma$ that assigns variables of \(\varphi_R\), the formula \(\varphi_R\{\sigma\}\) 
is obtained by replacing every variable by its corresponding constant or truth value.
If $\sigma$ assigns every variable of $\varphi$, then \(\varphi_R\{\sigma\}\) is ground and can be evaluated.
We will use the notation $\lf(\varphi_R\{\sigma\})$ for this evaluation.

\subsubsection{Aggregation and reification}

\new{
Essential to performing neurosymbolic computation is aggregation over labelled formulas.
The labelled formulas decompose in logical and neural labels in the unifying definition (Section \ref{sec:unification}).
In contrast, labelled formulas in \dl follow complex algebraic compositions, but \dl does not yet have a mechanism for aggregation.
Hence, the definition of a \dl formula and its labelling is extended to support aggregation functions \( g: \mathcal{A}_R^* \to \mathcal{A}_R \) \citep{grabisch2009aggregation}. 
}

\begin{definition}[Formula continued - aggregation]
    \label{def:aggregation}
    If \(\varphi_R\) is a formula that contains variable \(V\), and \(\boxplus: \mathcal{A}_R^* \to \mathcal{A}_R\) is an aggregating function, then \(\underset{V} \bigboxplus \varphi_R \) is also a formula.
\end{definition}

 \begin{definition}[Labelling continued - aggregation]
    Let  \(\underset{V} \bigboxplus \varphi_R \) be a formula and \(V\) a variable with domain \(dom(V)\).
    The label of the formula is an aggregation over the possible value assignments to \(V\) by applying \(\boxplus\), i.e.
   \begin{equation}
        \lf(\underset{V} \bigboxplus \varphi_R) = 
         \underset{c_i \in dom(V)} \bigboxplus \, \lf(\varphi_R\{V \mapsto c_i\}).
    \end{equation}
\end{definition}

\new{
Aggregation in \dl generalises quantification in first-order logic.
For instance, existential quantification over a formula labelled in the Boolean algebra $\mathbb{B}$ corresponds to using the infimum $\inf$ as aggregation function.
Typical other aggregations will be simple, like a summation over discrete variables or integration over continuous variables.
}

\begin{example}{\sffamily Aggregation over variables}
Consider the formula $\varphi_\mathbb{B}$ defined below. It uses an aggregation to express that all objects, which is represented as a regular variable $Object$ with $dom(Object) = \{box, triangle, square\}$, have a red color.
    \begin{equation}
        \varphi_\mathbb{B} = \underset{Object} \bigwedge \texttt{hasColor(Object,red)}_\mathbb{B}
    \end{equation}
\end{example}

\new{
\dl uses \emph{reification variables} to facilitate aggregation over truth values of atoms.
Aggregations in \dl (Definition~\ref{def:aggregation}) range over the domain of a variable.
However, many computational tasks of interest aggregate over assignments of truth values to atoms, e.g. AMC.
Assigning values to atoms instead of variables traditionally requires a language similar to second-order logic and so far \dl is more akin to first-order logic. The required second-order expressiveness is simulated in \dl using reification variables;
a special type of variables attached to algebraic atoms that represents the truth value of the atoms.
Reification variables are associated with an algebraic structure, from which they inherit their domain.
Reification and reified atoms are expressed using square brackets and extend simple algebraic atoms (Definition~\ref{def:simple-atom}).
}

\begin{definition}[Algebraic atom]
An algebraic atom is an expression of the form:
\[
    \texttt{p}(t_1,\dots,t_n)[X]_{R},
\]
where $\texttt{p}$ is an n-ary predicate, every $t_i$ is a constant or a regular variable, $R$ is an algebraic structure, and $X$ is either a reification variable or a constant truth value (typically a Boolean value or a fuzzy score).
If $[\cdot]$ is not present, then the algebraic atom is \emph{simple}.
\end{definition}

Reification variables otherwise behave exactly like regular variables; they can be assigned a value using a variable assignment and, crucially, can be aggregated over.
Aggregation over reification variables is necessary for modelling computational tasks such as algebraic model counting~\citep{kimmig_algebraic_2017} or weighted model integration~\citep{wmi}.

\begin{definition}[\new{Labelling of reified algebraic atom}]
Given a reified algebraic atom \(\texttt{p}(t_1,\dots,t_n)[X]_R\), with variable assignment \(\{X \mapsto x, \dots \} \), the labelling is denoted as
\[
    \lf(\texttt{p}(t_1,\dots,t_n)[x]_R).
\]
When the algebraic structure of $X$ is equal to $R$, then
\(
    \lf(\texttt{p}(t_1,\dots,t_n)[x]_R) = x,
\)
otherwise, the label is user-defined.
\end{definition}

\begin{example}{\sffamily Variable assignment}
    Consider the formula \(\varphi_\mathbb{P}\) that is defined below, where \(\mathbb{P}\) refers to the probabilistic structure that contains the arithmetic multiplication \(\times\) as a binary operation over the domain of real values. This formula contains two reification variables, $B$ and $E$.
    Depending on the truth value assigned to them, the label of $\varphi_\mathbb{P}$ changes.
    \begin{equation}
        \varphi_\mathbb{P} = (\texttt{burglary}[B]_\mathbb{P} \times_\mathbb{P} (\texttt{earthquake}[E]_\mathbb{P})
    \end{equation}
    The label of \(\varphi_\mathbb{P}\) depends on the labels of its ground reified algebraic atoms. As example, we consider the following definition.
    \begin{align*}
        \lf(\texttt{burglary}[true]_\mathbb{P}) &= 0.7 
        &\lf(\texttt{burglary}[false]_\mathbb{P}) &= 0.3 \\
        \lf(\texttt{earthquake}[true]_\mathbb{P}) &= 0.01 
        &\lf(\texttt{earthquake}[false]_\mathbb{P}) &= 0.99
    \end{align*}
    Based on this and the semantics of \(\times_\mathbb{P}\), it follows that the label of formula \(\varphi_\mathbb{P}[\sigma]\) corresponds to
    \begin{align*}
        \lf(\varphi_\mathbb{P}\{B \mapsto true, E \mapsto true\}) &= 0.7 \times 0.01, \\
        \lf(\varphi_\mathbb{P}\{B \mapsto true, E \mapsto false\}) &= 0.7 \times 0.99, \\
        \lf(\varphi_\mathbb{P}\{B \mapsto false, E \mapsto true\}) &= 0.3 \times 0.01, \\
        \lf(\varphi_\mathbb{P}\{B \mapsto false, E \mapsto false\}) &= 0.3 \times 0.99.
    \end{align*}
    \new{If $\texttt{earthquake}[E]_{\mathbb{P}}$ had instead been $\texttt{earthquake}[E]_{\mathbb{B}}$, then the algebraic structure of the algebraic atom would have been equivalent to that of reification variable $E$, i.e., structure $\mathbb{B}$. In such a case, the label would have been automatically defined \(\lf(\texttt{earthquake}[true]_\mathbb{B}) = true \) and \(\lf(\texttt{earthquake}[false]_\mathbb{B}) = false \).}
\end{example}

\begin{example}{\sffamily Variable assignments and neural labels}
    Before illustrating aggregation, we also demonstrate the use of regular variables.
    \[
        \varphi_\mathbb{B} = \texttt{burglary}(Video)[B]_\mathbb{B} \lor_\mathbb{B} \texttt{earthquake}(Seismic)[E]_\mathbb{B}
    \]
    This formula contains variables $B$, $E$, $Video$, and $Seismic$.
    If we apply the following variable assignment \(\sigma = \{B \mapsto true, E \mapsto true, Video \mapsto t_1, Seismic \mapsto t_2\} \), where \(t_1\) and \(t_2\) are data tensors, we get the ground formula:
    \[
        \varphi_\mathbb{B}\{\sigma\} = \texttt{burglary}(t1)[true]_\mathbb{B}\lor_\mathbb{B} \texttt{earthquake}(t2)[true]_\mathbb{B}.
    \]
    The labelling \(\lf\) is defined on a ground level. For example, with $nn$ a neural network that returns a probability,
    \begin{equation*}
        \lf(\texttt{burglary}(t_1)[true]_\mathbb{P}) = nn(t_1); \quad \quad \\
        \lf(\texttt{burglary}(t_1)[false]_\mathbb{P}) = 1-nn(t_1).
    \end{equation*}
    
    The labelling \(\lf\) must be defined for every possible valid variable assignment. When the domain is large, it is convenient to define them in a way that is applicable to multiple variable assignments. For example,
    \begin{equation*}
        \lf(\texttt{burglary}(Video)[B]_\mathbb{P}) = 
        \begin{cases}
            nn(Video) \ &\text{ if } B \qequal true, \\
            1-nn(Video) \ &\text{ otherwise }
        \end{cases}
    \end{equation*}
    This is merely notational as, again, the semantics are defined on the ground level.
\end{example}

\begin{example}{\sffamily Aggregation over reification variables}
    Let us extend one of the previous example formulae \(\varphi_\mathbb{P}\) with two aggregations using the standard summation operation \(\sum\).
    \begin{align*}
        \varphi_\mathbb{P} &= (\texttt{burglary}[B]_\mathbb{P} \times_\mathbb{P} \texttt{earthquake}[E]_\mathbb{P}) \\
        \varphi'_\mathbb{P} &= \sum_{B} \sum_{E} \varphi_\mathbb{P}
    \end{align*}
    By aggregating over variables $B$ and $E$ we change the resulting ground algebraic atom and thus the label. The labelling definition remains the same as before.
    \begin{align*}
        \lf(\texttt{burglary}[true]_\mathbb{P}) &= 0.7 
        &\lf(\texttt{burglary}[false]_\mathbb{P}) &= 0.3 \\
        \lf(\texttt{earthquake}[true]_\mathbb{P}) &= 0.01 
        &\lf(\texttt{earthquake}[false]_\mathbb{P}) &= 0.99
    \end{align*}
    
    Based on this, it follows that the label of formula \(\varphi'_\mathbb{P}\) corresponds to the equation below.
    \begin{align*}
        \lf(\varphi'_\mathbb{P}) 
        =\
        &\lf(\varphi_\mathbb{P}\{B \mapsto true, E \mapsto true\}) + 
        \lf(\varphi_\mathbb{P}\{B \mapsto true, E \mapsto false\}) \\
        +\
        &\lf(\varphi_\mathbb{P}\{B \mapsto false, E \mapsto true\}) + 
        \lf(\varphi_\mathbb{P}\{B \mapsto false, E \mapsto false\})
    \end{align*}
    \begin{equation*}
        \lf(\varphi'_\mathbb{P}) = (0.7 \times 0.01) + (0.7 \times 0.99) + (0.3 \times 0.01) + (0.3 \times 0.99) = 1.0
    \end{equation*}
\end{example}

\subsubsection*{Labels of atoms as parameters}
The labels associated to algebraic atoms are \textit{user-defined} inputs to the system. These labels play a role analogous to parameters in probabilistic models: they serve as the  parameters of the underlying DeepLog model. Because semantics are specified at the ground level, each ground algebraic atom must be externally assigned an appropriate label, i.e. a corresponding model parameter. In a learning setting, these parameters can be optimized as part of the learning process, typically guided by a loss function defined over the label of a formula (see Section \ref{sec:learning}).

For propositional algebraic atoms, i.e. zero-arity terms such as $burglary$, a label must be explicitly defined and supplied to the system. However, for structured atoms with arity greater than zero, e.g. $y\texttt{burglary}(t_1)$ or $\texttt{burglary}(B)$, we can avoid specifying labels for every possible ground term. Instead, we can use compact parameterizations that \textit{generalize} across argument substitutions for the same predicate. For instance, as illustrated in earlier examples, we can parameterize all atoms of the form $\texttt{burglary}(B)$ using a single neural network. This network leverages the tensor representations of constants that are (or will be) substituted for $B$, enabling the system to generalize across a potentially large domain.

The use of a continuous tensor space to provide a compact parameterization over a large domain of atoms is a central innovation in neurosymbolic systems. This approach contrasts with traditional fully-symbolic statistical relational AI frameworks~\citep{starai_book}, which typically support only all-or-nothing forms of generalization. From this perspective \citep{marra2024statistical}, DeepLog can be interpreted as a \textit{neural reparameterization} of statistical relational AI languages, where explicit parameters for ground algebraic atoms are replaced with differentiable representations that support smooth, data-driven generalization.

\subsubsection{Transformations between algebraic structure} 

A final component missing from \dl to model neurosymbolic computational tasks is the ability to work with more than one algebraic structure.
For instance, the weighted model count (WMC) of a sentence $\varphi$ aggregates the models, i.e. assignments to reification variables, that satisfy $\varphi$ in $\mathbb{B}$ multiplied by their probability in $\mathbb{P}$.
Hence, \dl is provided with transformations \( T_{R \to S}: \mathcal{A}_R \to \mathcal{A}_S \) between algebraic structures similar to the transformations of algebraic circuits~\citep{wang2024compositional}.

\begin{definition}[Formula continued - transformations]
If \( \varphi_R \) is a formula over $R$ and \(S\) an algebraic structure, then \( (\varphi_R)_S \) is a formula over \( S \).
\end{definition}

\begin{definition}[Labelling continued - transformations]
Let \( (\varphi_R)_S \) be a formula and \( T_{R \to S}: \mathcal{A}_R \to \mathcal{A}_S \) a function, then the label of \( (\varphi_R)_S \) is the transformation of the image of the label of \( \varphi_R \), i.e.
\[
\alpha((\varphi_R)_S) = T_{R \to S}( \alpha(\varphi_R) ).
\]
\end{definition}

Transformations together with aggregation over reification variables are sufficient to model weighted model counting and other common neurosymbolic computational tasks such as AMC, WMI, and fuzzy logic inference (Appendix~\ref{app:fuzzy_rewrite}).

\begin{example}{\sffamily Transformations - weighted model count}
    Let us once more adapt the earthquake-burglary example formula \(\varphi_\mathbb{P}\).
    \begin{align*}
        \varphi_\mathbb{P} &= \sum_{B} \sum_{E} (\texttt{burglary}[B]_\mathbb{B} \lor_\mathbb{B} \texttt{earthquake}[E]_\mathbb{B})_\mathbb{P} \times_\mathbb{P} (\texttt{burglary}[B]_\mathbb{P} \times_\mathbb{P} \texttt{earthquake}[E]_\mathbb{P})
    \end{align*}
    The logic part of this formula indicates our interest in having either \(\texttt{burglary}[true]\) or \(\texttt{earthquake}[true]\). The remaining parts represents the probability of that situation.
    \begin{align*}
        \lf(\texttt{burglary}[true]_\mathbb{B}) &= true 
        &\lf(\texttt{burglary}[false]_\mathbb{B}) &= false \\
        \lf(\texttt{earthquake}[true]_\mathbb{B}) &= true 
        &\lf(\texttt{earthquake}[false]_\mathbb{B}) &= false \\
        \lf(\texttt{burglary}[true]_\mathbb{P}) &= 0.7 
        &\lf(\texttt{burglary}[false]_\mathbb{P}) &= 0.3 \\
        \lf(\texttt{earthquake}[true]_\mathbb{P}) &= 0.01 
        &\lf(\texttt{earthquake}[false]_\mathbb{P}) &= 0.99
    \end{align*}
Finally, as transformation \(T_{\mathbb{B} \to \mathbb{P}}(x)\) we use the Iverson function \(\ive{\cdot}\), which maps \(true\) to \(1\), and \(false\) to \(0\). From this, it follows that the label of formula \(\varphi_\mathbb{P}\) corresponds to
    \begin{align*}
        \lf(\varphi_\mathbb{P}) 
        = 
        &(\ive{true} \times 0.7 \times 0.01) + 
        (\ive{true} \times 0.7 \times 0.99) + \\
        &(\ive{true} \times 0.3 \times 0.01) + 
        (\ive{false} \times 0.3 \times 0.99) = 0.703 
    \end{align*}
This coincides with the definition of the Weighted Model Count~\citep{Derkinderen24Semiring}.
\end{example}

\subsection{Neurosymbolic DeepLog Model}

We are now able to formally define 
a DeepLog model:
\begin{definition}[DeepLog Model]
    A DeepLog model consists of the following components:
    \begin{itemize}
    \item a set of variables $\mathcal{V}$, \new{their domain (for each regular variable), and their corresponding algebraic structure (for each reification variable),}
    \item a set of algebraic structures ${\cal SA}$,
    \item a set of transformation functions ${ST}$,
    \item a DeepLog formula \(\varphi\) using algebraic operators from algebras in ${\cal SA}$,
    transformations in $ST$,
    \item and a labelling \(\lf\) for \(\varphi\).
\end{itemize}
Given a DeepLog model with formula \(\varphi\), and a variable assignment \(\sigma\) to all unassigned variables \(\mathbf{V} \subseteq \mathcal{V}\) in $\varphi$, a DeepLog model outputs the label \(\lf(\varphi\{\sigma\})\).  Computing this label constitutes the inference task of DeepLog.
\end{definition}

This is the central definition of DeepLog. It specifies the elements that DeepLog models consist of, as well as what inference for DeepLog means.

\begin{example}{\sffamily A DeepLog Model and Inference Task}

Reconsider the example from Section~\ref{sec:illustration}.
The corresponding DeepLog model is defined as follows, where we abbreviate \texttt{burglary} to \texttt{b}, \texttt{earthquake} to \texttt{e}, $Video$ to $V$ and $Seismic$ to $S$.

\begin{itemize}
    \item \(\mathcal{V} = \{B,E,V,S\}\)
    
    The domain \(dom(V)\) are tensors corresponding to video data, and $dom(S)$ are tensors corresponding to seismic data. Both \(B\) and \(E\) \new{are associated to the Boolean algebraic structure, meaning they are assigned Boolean truth values.}

    \item \({\cal SA} = \{ \mathbb{B} , \mathbb{P}\}\)
    
    As algebraic structures we consider   \(\mathbb{B}\) and \(\mathbb{P}\), the Boolean- and probabilistic semirings, \new{with aggregation $\sum$ the standard arithmetic addition operator that we use in $\mathbb{P}$.}

    \item  $ST =  \{
    T_{\mathbb{B} \rightarrow \mathbb{P}} \}$
    
    where  \(T_{\mathbb{B} \rightarrow \mathbb{P}} = \ive{\cdot}\), which is the Iverson function that maps \(true\) to \(1\) and \(false\) to \(0\).

    \item The labeling function $\alpha$:
    \begin{align*}
        \lf(\texttt{b}(V)[true]_\mathbb{B}) &= true 
        &\lf(\texttt{b}(V)[false]_\mathbb{B}) &= false \\
        \lf(\texttt{e}(S)[true]_\mathbb{B}) &= true 
        &\lf(\texttt{e}(S)[false]_\mathbb{B}) &= false \\
        \lf(\texttt{b}(V)[true]_\mathbb{P}) &= nn_b(V) 
        &\lf(\texttt{b}(V)[false]_\mathbb{P}) &= 1 - nn_b(V) \\
        \lf(\texttt{e}(S)[true]_\mathbb{P}) &= nn_e(S) 
        &\lf(\texttt{e}(S)[false]_\mathbb{P}) &= 1 - nn_e(S)
    \end{align*}

    \item The formula $\varphi_\mathbb{P}
    :=$ 
\begin{align*}
    \sum_{B} \sum_{E} (\texttt{b}(V)[B]_\mathbb{B} \lor_\mathbb{B}
        \texttt{e}(S)[E]_\mathbb{B})_\mathbb{P} \times_\mathbb{P}
        (\texttt{b}(V)[B]_\mathbb{P} \times_\mathbb{P} \texttt{e}(S)[E]_\mathbb{P})
\end{align*}
\end{itemize} 

Given a variable assignment $\sigma = \{V \mapsto \iimage{images/burglary1}, S \mapsto \iimage{images/earthquake1}\}$, the inference task over the previous model is computing the label:
\begin{align*}
\displaystyle 
\alpha\big(\varphi_\mathbb{P}[\sigma] \big)
        = 
        &(\ive{true} \times nn_b(\iimage{images/burglary1}) \times nn_e(\iimage{images/earthquake1})) + \\
        &
        (\ive{true} \times nn_b(\iimage{images/burglary1}) \times (1 - nn_e(\iimage{images/earthquake1}))) + \\
        &(\ive{true} \times (1 - nn_b(\iimage{images/burglary1})) \times nn_e(\iimage{images/earthquake1})) +\\
        &
        (\ive{false} \times (1 - nn_b(\iimage{images/burglary1})) \times (1 - nn_e(\iimage{images/earthquake1}))) 
\end{align*}
\end{example}

The above example shows a DeepLog model expressed at the intermediate level, which defines the semantics of the model and the necessary computation.
In  Section \ref{sec:implementation}, we will show how this intermediate level representation can be mapped onto the computation level by means of algebraic circuits and computational graphs (cf. Figures \ref{fig:intro_circuit_exa_full_repeat} and \ref{fig:intro_circuit_exa_opt_lb_repeat}).
\textbf{The combination of \dl models and their operationalisations as algebraic circuits constitute the DeepLog neurosymbolic machine.}

\subsection{Weakly supervised learning in DeepLog}
\label{sec:learning}

DeepLog  fits in a wide variety of learning problems.
This section demonstrates how to use DeepLog within the typical neurosymbolic AI learning setting of \emph{weakly supervised learning}.

In this setting, the objective is to train multiple neural networks, each responsible for predicting a different class, like \texttt{burglary} or \texttt{earthquake}.
Direct supervision for these networks is not available. Instead of explicit labels for each class, the training data provides only global labels that indicate whether certain logical relations among the classes are satisfied. 
As a result, the supervision does not specify which particular network should activate for a given input -- as in standard supervised learning --  but constrains them indirectly through the formulas. This renders the supervision inherently weak: the formulas act as constraints, providing information about joint outcomes without explicitly indicating the contribution of each individual network.

Therefore, we consider a dataset of annotated logical formulas; i.e.  
\(
\mathcal{D} = \{(\varphi_i, \sigma_i, y_i)\}_{i=1}^N
\).
Each data point consists of  a DeepLog formula $\varphi_i$,  a variable assignment $\sigma_i$, and a ground-truth label $y_i$ for $\varphi_i$. The formula $\varphi_i$ may be either \emph{ground}, in which case $\sigma_i$ is empty, or it may contain \emph{free variables}, in which case $\sigma_i$ specifies their values. Free variables, like the video \textit{V} or the sensor \textit{S}, are used to feed input tensors to the neural networks appearing in the formulas. Moreover, the formulas $\varphi_i$ are expected to aggregate over possible truth assignments, as we do not know which, and to what extent, each scenario contributed to the observed label $y_i$  (i.e. the weakly supervised setting). 
Finally, let $\theta$ denote the set of DeepLog parameters. These include either the labels of the ground atomic formulas or the weights of neural network that parameterize such labels.

Then, learning in DeepLog corresponds to the following optimization problem:
\[
\min_{\theta} \sum_{(\varphi_i, \sigma_i, y_i) \in \mathcal{D}} \ell\Big(\alpha(\varphi_i[\sigma_i]), y_i\Big)
\]
where $\ell(\cdot)$ is a user defined loss function\new{, and $\lf$ is parameterised by $\theta$}. For example in a probabilistic setting, where labels are interpreted as probabilities, $\ell$ can be instantiated as the binary cross-entropy loss.

\begin{example}{\sffamily Learning alarm }
    Consider the alarm example from before. We can build a weakly supervised learning setting where we have supervisions only on whether an alarm went off or not, but not its cause, i.e., either burglary or earthquake. Consider the previously introduced DeepLog formula:

    \[\varphi = \sum_{B} \sum_{E} (\texttt{b}(V)[B]_\mathbb{B} \lor_\mathbb{B}
        \texttt{e}(S)[E]_\mathbb{B})_\mathbb{P} \times_\mathbb{P}
        (\texttt{b}(V)[B]_\mathbb{P} \times_\mathbb{P} \texttt{e}(S)[E]_\mathbb{P})
    \]

    We can build a  dataset as: 
    \begin{align*}
    \mathcal{ D} = \{ & \big(\varphi,  \{V \mapsto \iimage{images/burglary1}, S \mapsto \iimage{images/earthquake1}\}, 0\big), \\
    & \big(\varphi,  \{V \mapsto \iimage{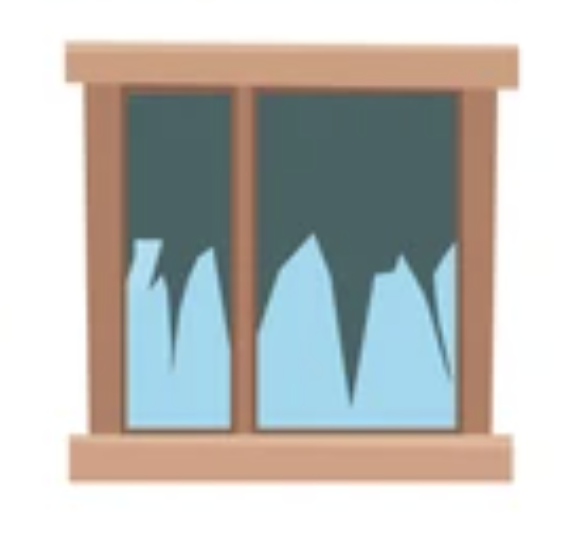}, S \mapsto \iimage{images/earthquake1}\}, 1\big),\\ & 
    \ldots\}
    \end{align*}
\end{example}

\section{Efficient Computation with Algebraic Circuits}\label{sec:implementation}

The success of deep learning has been 
enabled by the availability of flexible, efficient and open-source software packages that are compatible with modern hardware accelerators such as GPUs.
The same compatibility with modern hardware accelerators as well as efficient algorithms are necessary for neurosymbolic methods to be adopted.
It motivates a discussion on (tractable) representations for computing labels of formulae.

Algebraic circuits are a ubiquitous representation for tractable computation.
They are actively used and developed in the field of statistical relational AI in the form of arithmetic circuits~\citep{darwiche2003differential}, probabilistic circuits~\citep{vergari2021compositional}, sum product networks~\citep{poon2011sum}, etc. as they facilitate tractable probabilistic inference (e.g. computing marginal probabilities).
This tractability has succesfully lead to applications of algebraic circuits in neurosymbolic AI~\citep{manhaeve2018deepproblog}, where they serve as computational graphs.


\subsection{Algebraic Circuits}

An \emph{algebraic circuit} is represented as a directed acyclic computational graph whose behavior is associated with an algebraic structure~\citep{Ruiz2021Circuits}. 
We focus on algebraic circuits that compose labels using only two binary operations, $\oplus$ and $\otimes$.
In contrast to prior works in probabilistic logic, we do not assume that those operations form a semiring per se,  although this property is desirable, as we discuss in more detail later on. 
To define an algebraic circuit, we use the concept of a labelling function: a function that acts as a labelling \(\lf\) for a specific formula \(\varphi\).
\begin{definition}[Labelling function]\label{def:labelling_function}
    A labelling function \(\lf_{\varphi_R}\) is a function that takes as argument a variable assignment \(\sigma\), and returns the label of \(\varphi_R\), i.e. \(\lf_{\varphi_R}(\sigma) = \lf(\varphi_R\{\sigma\})\).
    Its co-domain is the set of labels $\mathcal{A}_R$ of the algebraic structure $R$.
\end{definition}

\begin{definition}[Algebraic Circuit]
   Given a set of labelling functions $\mathcal{F}$ with co-domain $\mathcal{A}_R$, and binary operations $\oplus$ and $\otimes$ over $\mathcal{A}_R$, an algebraic circuit $C = (\mathcal{G}, \mathcal{F}, \oplus, \otimes)$ is a directed acyclic graph $\mathcal{G}$  where each leaf node $l$ is associated with a labelling function $\lf_l \in \mathcal{F}$, and each internal node is associated with either $\oplus$ or $\otimes$. 
   Each root node $n$ of the algebraic circuit represents a labelling function that is inductively defined as
    \begin{equation*}
        \lf_n(\sigma) \defeq \begin{cases}
                    \lf_l(\sigma) \quad &\text{ if $n$ is a leaf node $l$} \\
                    \oplus^k_{i=1} \lf_{n_i}(\sigma) \quad &\text{ if } n = \oplus \\
                    \otimes^k_{i=1} \lf_{n_i}(\sigma) \quad &\text{ if } n = \otimes
                    \end{cases}\quad,
    \end{equation*}
    with each $n_i$ a child node of internal node $n$, \(\sigma\) a variable assignment.
\end{definition}


By limiting an algebraic circuit to two binary operations, we simplify its theoretical study and the design of new optimisation techniques. As a very simple example, consider idempotency as a property. If $\oplus$ is a logical disjunction operation $\lor$, then idempotency causes $f(\mathbf{V}) \oplus f(\mathbf{V}) = f(\mathbf{V})$, so we can reduce the computational graph in case $f(\mathbf{V}) \oplus f(\mathbf{V})$ occurs.
Typically, we will consider two operations that form a semiring, as their theoretical properties enable several interesting optimizations, allowing us to rewrite and optimize larger parts of the formula. We refer to \citet{KimmigBR17AMC,wang2024compositional} for more details, but will provide intuition later on with the example of probabilistic logic.

\begin{example}{\sffamily Algebraic circuits: propositional logic circuits}
    \new{Consider as example the Boolean function $a \lor (\neg a \land b)$, which we express as DeepLog formula:}
    \[
        \texttt{a}[A]_\mathbb{B} \vee_\mathbb{B} (\neg_\mathbb{B} \texttt{a}[A]_\mathbb{B} \wedge_\mathbb{B} \texttt{b}[B]_\mathbb{B}).
    \]

    The following algebraic circuit $C$ represents this formula by combining three labelling functions $\lf_a$, $\lf_{\neg a}$, and $\lf_b$, by using for $\oplus$ the logical disjunction \(\lor\), and for $\otimes$ the logical conjunction \(\land\). 

    \begin{center}
        \begin{minipage}{.4\linewidth}
            \centering
            \begin{tikzpicture} [node distance=0.9cm]

    \node (n1) [plus] {$\oplus$};
    \node (na) [leaf, below of=n1, left of=n1, xshift=0.2cm] {$\lf_a$};
    \node (n2) [times, below of=n1, right of=n1, xshift=-0.2cm] {$\otimes$};

    \node (nna) [leaf, below of=n2, left of=n2, xshift=0.2cm] {$\lf_{\neg a}$};
    \node (nb) [leaf, below of=n2, right of=n2, xshift=-0.2cm] {$\lf_{b}$};

    \draw [dirarrow] (na) -- (n1);
    \draw [dirarrow] (n2) -- (n1);
    \draw [dirarrow] (nna) -- (n2);
    \draw [dirarrow] (nb) -- (n2);
\end{tikzpicture}
  
        \end{minipage}
        \begin{minipage}{.5\linewidth}
            \begin{minipage}{.2\linewidth}
                where
            \end{minipage}
            \begin{minipage}{.3\linewidth}
                \begin{align*}
                    \lf_a(\sigma) &=                     \begin{cases}
                        true &\text{ if } A \qequal true \text{ in } \sigma \\
                        false &\text{ otherwise}
                    \end{cases} \\
                    \lf_{\neg a}(\sigma) &= 
                    \begin{cases}
                        false &\text{ if } A \qequal true \text{ in } \sigma \\
                        true &\text{ otherwise}
                    \end{cases} \\
                    \lf_{b}(\sigma) &= 
                    \begin{cases}
                        true &\text{ if } B \qequal true \text{ in } \sigma \\
                        false &\text{ otherwise}
                    \end{cases}
                \end{align*}
            \end{minipage}
        \end{minipage}
    \end{center}
    Naturally, each labelling function $\lf_i \in \mathcal{F}$ emits a Boolean value, $true$ or $false$. 
    The circuit forms a labelling function $\lf_C$ which can be evaluated. In this example, $\lf_C(\{A \mapsto false, B \mapsto false\}) = false$ while the other assignments evaluate to $true$.
\end{example}

\subsection{Composing Algebraic Circuits}\label{sec:composing_ac}

\begin{figure}
    \centering
    \begin{tikzpicture} [node distance=1cm]
    \node (n1) [leaf] {$\sum_{B} \sum_{E}$};
    \node (n2) [leaf, below of=n1] {$\times_\mathbb{P}$};
    \node (n3) [leaf, below of=n2, right of=n2, xshift=0.5cm] {$\times_\mathbb{P}$};
    \node (n4) [leaf, below of=n2, left of=n2, xshift=-0.5cm] {$T_{\mathbb{B} \rightarrow \mathbb{P}}$};
    \node (n5) [leaf, below of=n3, left of=n3] {$\lf_{b(V)[B]_\mathbb{P}}(\sigma)$};
    \node (n6) [leaf, below of=n3, right of=n3] {$\lf_{e(S)[E]_\mathbb{P}}(\sigma)$};
    \node (n7) [leaf, below of=n4] {$\lor_\mathbb{B}$};
    \node (n8) [leaf, below of=n7, left of=n7] {$\lf_{b(V)[B]_\mathbb{B}}(\sigma)$};
    \node (n9) [leaf, below of=n7, right of=n7] {$\lf_{e(S)[E]_\mathbb{B}}(\sigma)$};

    \draw [arrow] (n2) -- (n1);
    \draw [arrow] (n3) -- (n2);
    \draw [arrow] (n4) -- (n2);
    \draw [arrow] (n5) -- (n3);
    \draw [arrow] (n6) -- (n3);
    \draw [arrow] (n7) -- (n4);
    \draw [arrow] (n8) -- (n7);
    \draw [arrow] (n9) -- (n7);

    \draw[blue, thick, rounded corners] 
        ([xshift=-0.7cm]n4.north) -- 
        ([xshift=0.7cm]n4.north) -- 
        ([xshift=0.5cm]n7.east) -- 
        ([yshift=0.1cm]n9.north east) --
        ([yshift=-0.1cm]n9.south east) --
        ([yshift=-0.1cm]n8.south west) --
        ([yshift=0.1cm]n8.north west) --
        cycle;

    \draw[red, thick, rounded corners] 
        ([yshift=0.2cm,xshift=-0.25cm]n7.north) -- 
        ([yshift=0.2cm,xshift=0.25cm]n7.north) -- 
        ([yshift=0.2cm,xshift=0.25cm]n7.south east) -- 
        ([xshift=-0.1cm]n9.north east) --
        ([xshift=-0.1cm]n9.south east) --
        ([xshift=0.1cm]n8.south west) --
        ([xshift=0.1cm]n8.north west) --
        cycle;

    \draw[green, thick, rounded corners] 
        ([xshift=-0.8cm]n4.north) --
        ([yshift=0.2cm,xshift=-0.2cm]n2.north) --
        ([yshift=0.2cm,xshift=0.2cm]n2.north) --
        ([xshift=0.1cm]n6.north east) --
        ([xshift=0.1cm]n6.south east) --
        ([xshift=3.1cm, yshift=-0.2cm]n9.south east) --
        ([xshift=0.3cm, yshift=-0.2cm]n9.south east) --
        ([xshift=-0.1cm, yshift=-0.2cm]n8.south west) --
        ([xshift=-0.1cm, yshift=.2cm]n8.north west) --
        cycle;

\end{tikzpicture}
  
    \caption{The algebraic circuit corresponding to the expectation that a burglary or earthquake happened. The colored shape around a subset of the computational graph indicates the algebraic circuit composition.}
    \label{fig:intro_circuit_exa_full_repeat}
\end{figure}
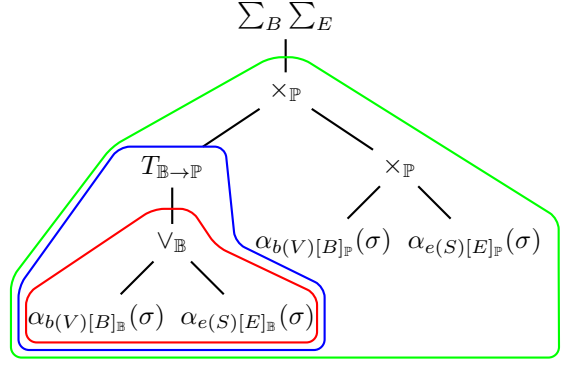

Algebraic circuits are composable computational graphs. This means we can take two circuits and an operator and build a combined one. It also
 means that even formulae with more than two binary operations can be represented using an algebraic circuit---one that is a modular combination of different algebraic circuits~\citep{wang2024compositional}. 
For example, the formula that represents the expectation of a burglary or earthquake happening is representable as an algebraic circuit, yet it contains more than just two types of binary operations.

\begin{equation*}
    \sum_{B} \sum_{E} (\texttt{b}(V)[B]_\mathbb{B} \lor_\mathbb{B}
        \texttt{e}(S)[E]_\mathbb{B})_\mathbb{P} \times_\mathbb{P}
        (\texttt{b}(V)[B]_\mathbb{P} \times_\mathbb{P} \texttt{e}(S)[E]_\mathbb{P})
\end{equation*}
Figure~\ref{fig:intro_circuit_exa_full_repeat} illustrates this: the nodes that are delineated by the red shape form a single algebraic circuit, that can still be used in the implementation of the blue-delineated algebraic circuit, which in turn can be used to form the larger green-delineated algebraic circuit.

This compositional structure is what grants DeepLog modularity, enabling the construction of novel neurosymbolic frameworks on top of basic labeling functions (e.g., Boolean, fuzzy, probabilistic). Furthermore, once defined, these higher-level frameworks can themselves be reused as building blocks, aiming at the realisation of a rich ecosystem of interoperable components.

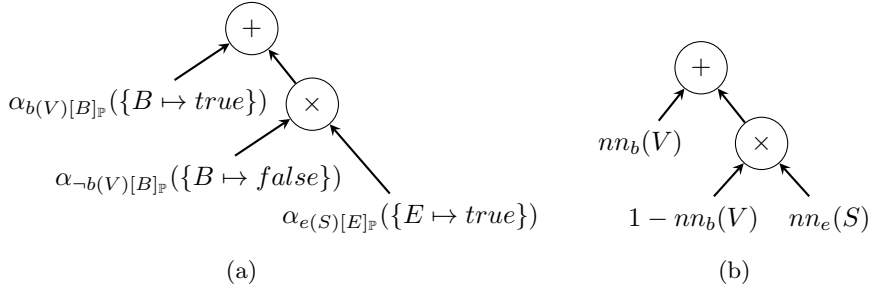
\begin{figure}
    \centering
    \begin{subfigure}{.4\linewidth}
        \centering
        \begin{tikzpicture} [node distance=1cm]
    \node (n1) [plus] {$+$};
    \node (na) [leaf, below of=n1, left of=n1, xshift=-0.5cm] {$\lf_{b(V)[B]_\mathbb{P}}(\{B \mapsto true\})$};
    \node (n2) [times, below of=n1, right of=n1, xshift=-0.2cm] {$\times$};

    \node (nna) [leaf, below of=n2, left of=n2, xshift=-0.5cm] {$\lf_{\neg b(V)[B]_\mathbb{P}}(\{B \mapsto false\})$};
    \node (nb) [leaf, below of=n2, right of=n2, xshift=0.3cm, yshift=-0.5cm] {$\lf_{e(S)[E]_\mathbb{P}}(\{E \mapsto true\})$};

    \draw [dirarrow] (na) -- (n1);
    \draw [dirarrow] (n2) -- (n1);
    \draw [dirarrow] (nna) -- (n2);
    \draw [dirarrow] (nb) -- (n2);
\end{tikzpicture}
  
        \caption{}
        \label{fig:intro_circuit_exa_opt_lb_repeat}
    \end{subfigure}
    \begin{subfigure}{.4\linewidth}
        \centering
         \begin{tikzpicture} [node distance=1cm]

    \node (n1) [plus] {$+$};
    \node (na) [leaf, below of=n1, left of=n1, xshift=0.2cm] {$nn_b(V)$};
    \node (n2) [times, below of=n1, right of=n1, xshift=-0.2cm] {$\times$};

    \node (nna) [leaf, below of=n2, left of=n2, xshift=0.1cm] {$1 - nn_b(V)$};
    \node (nb) [leaf, below of=n2, right of=n2, xshift=-0.1cm] {$nn_e(S)$};

    \draw [dirarrow] (na) -- (n1);
    \draw [dirarrow] (n2) -- (n1);
    \draw [dirarrow] (nna) -- (n2);
    \draw [dirarrow] (nb) -- (n2);
\end{tikzpicture}
  
         \caption{}
    \end{subfigure}
    \caption{An algebraic circuit that is semantically equivalent to Figure~\ref{fig:intro_circuit_exa_full_repeat}. The right-side version is the same circuit, but with the labelling functions \(\lf\) replaced by their definition.}
    \label{fig:circuit-exa}
\end{figure}

\subsection{Rewriting Algebraic Circuits for Optimization}

The structural properties of algebraic structures may enable the rewriting of a formula into a semantically equivalent one that leads to more efficient or more simple algebraic circuits.
To provide intuition on this, we will show how to rewrite the formula for the neurosymbolic probabilistic logic setting into a single algebraic circuit (cf. Figure~\ref{fig:intro_circuit_exa_opt_lb_repeat}). {Afterwards, we also discuss the fuzzy settings.} 

\new{It is clear how to rewrite these settings, and any instances of them can be directly expressed in their final more optimized form. An automatic rewriting of general formulae, however, is an interesting open research challenge.}

\subsubsection{Algebraic Circuits in Neurosymbolic Probabilistic Logic}
\label{subsec:circuit_rewrite:dpl}
The formula of a neurosymbolic probabilistic logic setting corresponds to
\begin{equation*}
    \sum_{A_1} \ldots \sum_{A_n} \;
        (\varphi_\mathbb{B})_\mathbb{P} \times_\mathbb{P}
        \varphi'_\mathbb{P},
\end{equation*}
with \(\varphi_\mathbb{B}\) a formula that evaluates to a Boolean label and that is comprised of algebraic atoms $\texttt{a}_1(V)[A_1]_\mathbb{B}$, $\dots$, $\texttt{a}_n(V)[A_n]_\mathbb{B}$, 
and \(\varphi'_\mathbb{P}\) a formula comprising algebraic atoms \(\texttt{a}_1(V)[A_1]_\mathbb{P}, \dots, \texttt{a}_n(V)[A_n]_\mathbb{P}\), representing the probability of the interpretation $\{\texttt{a}_1(V):=A_1, \ldots, \texttt{a}_n(V) := A_n\}$. The full formula corresponds to the expectation of satisfying \(\varphi_\mathbb{B}\).

A popular approach to compute this efficiently---an approach also used by DeepProbLog---is to assume a factorized weight function $\varphi'_\mathbb{P}$. For instance, the previously used example of burglary and earthquake, which we repeat below, is an example of a neurosymbolic probabilistic logic setting where we assume a factorization of the probability function
\begin{align}\label{eq:functional_dpl_example} 
    \sum_{B} \sum_{E} (\texttt{b}(V)[B]_\mathbb{B} \lor_\mathbb{B}
        \texttt{e}(S)[E]_\mathbb{B})_\mathbb{P} \times_\mathbb{P}
        (\texttt{b}(V)[B]_\mathbb{P} \times_\mathbb{P} \texttt{e}(S)[E]_\mathbb{P}).
\end{align}
The circuit corresponding to this formula is shown in Figure~\ref{fig:intro_circuit_exa_full_repeat}. Importantly, the assumption that \(\varphi_\mathbb{P}\) factorizes, preferably into \(\prod_{i} a_i[V_i]_\mathbb{P}\), does not limit the expressivity in the context of Boolean variables because we can introduce additional variables and model dependencies through the logical component \(\varphi_\mathbb{B}\)~\citep{ChaviraD08,Derkinderen24Semiring}. Appendix~\ref{app:dependencies} illustrates this through an example.

With a factorized probability function, the example formula can be reformulated into an equivalent but more efficient form (cf. Figure~\ref{fig:circuit-exa}). To achieve this, we first reformulate the logical component \(\texttt{b}(V)[B]_\mathbb{B} \lor_\mathbb{B} \texttt{e}(S)[E]_\mathbb{B}\) into \(
\texttt{b}(V)[B]_\mathbb{B} \lor_\mathbb{B}
        \big( 
            \texttt{e}(S)[E]_\mathbb{B} \land_\mathbb{B} 
            \neg_\mathbb{B} \texttt{b}(V)[B]_\mathbb{B} 
        \big)
\).
This logical formula is said to be in \emph{deterministic, decomposable negation normal form} (d-DNNF). We omit the details of this form~\citep{darwiche2002knowledge}, but it is the properties of this form paired with the homomorphism property of the transformation that allows us to push down the transformation operation. More specifically, to easily push the transformation \(T_{\mathbb{B} \rightarrow \mathbb{P}}\) into the leaf nodes where it becomes part of the leaf node's function, we rely on the following two properties:
\begin{equation*}
    (\varphi_{\mathbb{B},1} \land_\mathbb{B} \varphi_{\mathbb{B},2})_\mathbb{P} 
        = (\varphi_{\mathbb{B},1})_\mathbb{P} \times (\varphi_{\mathbb{B},2})_\mathbb{P} 
    \qquad \text{ and } \qquad 
    (\varphi_{\mathbb{B},1} \lor_\mathbb{B} \varphi_{\mathbb{B},2})_\mathbb{P} = 
        (\varphi_{\mathbb{B},1})_\mathbb{P} + (\varphi_{\mathbb{B},2})_\mathbb{P}.
\end{equation*}
The latter property is only true iff \(\varphi_{\mathbb{B},1}\) and \(\varphi_{\mathbb{B},2}\) do not both evaluate to true for any given interpretation---the determinism property~\citep{darwiche2002knowledge}. 
In case the logical component \(\varphi_\mathbb{B}\) is not a d-DNNF, knowledge compilation techniques can be used to automatically transform it into this form~\citep{pysdd,Lagniez17D4}.
For more details on the connection between the d-DNNF properties and inference, we refer to \citet{KimmigBR17AMC}.

After pushing the transformation into the leaf nodes, we obtain
\begin{equation}
    \sum_{B} \sum_{E} 
        \Big( 
            (\texttt{b}(V)[B]_\mathbb{B})_\mathbb{P} +_\mathbb{P} 
                \big(
                    (\texttt{e}(S)[E]_\mathbb{B})_\mathbb{P} \times_\mathbb{P} (\neg_\mathbb{B} \texttt{b}(V)[B]_\mathbb{B})_\mathbb{P}
                \big)
        \Big)
        \times_\mathbb{P}
        \Big( \texttt{b}(V)[B]_\mathbb{P} \times_\mathbb{P} \texttt{e}(S)[E]\_\mathbb{P} \Big).
\end{equation}

Next, we push down the probabilistic component \(\texttt{b}(V)[B]_\mathbb{P} \times_\mathbb{P} \texttt{e}(S)[E]_\mathbb{P}\) by exploiting distributivity.

\begin{equation}
\begin{aligned}
    \sum_{B} \sum_{E} 
    &\Bigg( 
    (\texttt{b}(V)[B]_\mathbb{B})_\mathbb{P} \times_\mathbb{P} \Big( \texttt{b}(V)[B]_\mathbb{P} \times_\mathbb{P} \texttt{e}(S)[E]_\mathbb{P} \Big) 
    \Bigg) 
    +_\mathbb{P} \\
    &\qquad
    \Bigg(
        \big(
                    (\texttt{e}(S)[E]_\mathbb{B})_\mathbb{P} \times_\mathbb{P} (\neg_\mathbb{B} \texttt{b}(V)[B]_\mathbb{B})_\mathbb{P}
        \big)
        \times_\mathbb{P}
        \Big( \texttt{b}(V)[B]_\mathbb{P} \times_\mathbb{P} \texttt{e}(S)[E]_\mathbb{P} \Big)
    \Bigg)
    \end{aligned}
\end{equation}
Then, we push down the aggregation operations. These will interact with the Boolean algebraic atoms that function as a filter, resolving the aggregation. For example, through the reification variable $B$, \((\texttt{b}(V)[B]_\mathbb{B})_\mathbb{P}\) effectively selects the interpretation wherein \(\texttt{b}(V)\) is true. To properly illustrate this, we use the operational notation. For example, we use \(\lf_{\texttt{b}(V)[B]_\mathbb{B}}(\sigma)\) instead of \(\texttt{b}(V)_\mathbb{B}[V]\). We repeat the previous formula, but now using the operational notation.
\begin{align*}
    \lf_\varphi(\sigma) = \Bigg(
    \sum_{B} \sum_{E}
    &\ive{\lf_{\texttt{b}(V)[B]_\mathbb{B}}(\sigma)} \times \lf_{\texttt{b}(V)[B]_\mathbb{P}}(\sigma) \times \lf_{\texttt{e}(S)[E]_\mathbb{P}}(\sigma)
    \Bigg) 
    + \\
    &\Bigg(
     \sum_{B} \sum_{E} 
        \big(
                    \ive{\lf_{\texttt{e}(S)[E]_\mathbb{B}}(\sigma)} \times \ive{\lf_{\neg_\mathbb{B} \texttt{b}(V)[B]_\mathbb{B}}(\sigma)}
        \big)
        \times
        \lf_{\texttt{b}(V)[B]_\mathbb{P}}(\sigma) \times \lf_{\texttt{e}(S)[E]_\mathbb{P}}(\sigma)
    \Bigg)
\end{align*}
By resolving the aggregation operations, and by exploiting commutativity, associativity, and the fact that \(\lf_{\texttt{b}(V)[B]_\mathbb{P}}\) is independent of \(E\), we obtain the equation below.
\begin{align*}
    \lf_\varphi(\sigma) = \Bigg(
    \lf_{\texttt{b}(V)[true]_\mathbb{P}}(\sigma) \times \sum_{E} \lf_{\texttt{e}(S)[E]_\mathbb{P}}(\sigma)
    \Bigg) 
    + 
    \Bigg( 
        \lf_{\texttt{b}(V)[false]_\mathbb{P}}(\sigma) \times \lf_{\texttt{e}(S)[true]_\mathbb{P}}(\sigma)
    \Bigg)
\end{align*}
Finally, we exploit the following facts: \(\lf_{\texttt{e}(S)[E]_\mathbb{P}}\) is independent of \(B\), 
\(\lf_{\texttt{b}(V)[B]_\mathbb{P}}\) is independent of \(E\), \(\sum_{E} \lf_{\texttt{e}(S)[E]_\mathbb{P}}(\sigma)  = nn_e(S) + (1 - nn_e(S)) = 1.0\), and $a \times 1 = a$. This yields the equation below.
\begin{align*}
    \lf_\varphi(\sigma) = \Bigg(
    \lf_{\texttt{b}(V)[true]_\mathbb{P}}(\sigma)
    \Bigg) 
    +
    \Bigg( 
        \lf_{\texttt{b}(V)[false]_\mathbb{P}}(\sigma) \times \lf_{\texttt{e}(S)[true]_\mathbb{P}}(\sigma)
    \Bigg)
\end{align*}
The resulting equation consists of three labelling functions, each of which represents a single probability value. In the neurosymbolic setting this becomes 
\begin{align*}
    \lf_\varphi(\sigma) = nn_b(V) + (1-nn_b(V)) \times nn_e(S),
\end{align*}
which corresponds to the circuit in Figure~\ref{fig:circuit-exa}.

We have now illustrated how a DeepLog model can be manipulated and  optimized in order to obtain a simple and efficient algebraic circuit.

\subsubsection{Algebraic Circuits in Neurosymbolic Fuzzy Logic}

The knowledge compilation step that is performed while rewriting the probabilistic logic formula can be computationally expensive, even when performed offline. To avoid this step, some works have explored replacing the standard Boolean logic semantics with fuzzy semantics. In practice, this typically means that the otherwise Boolean operations of a logic formula are replaced with fuzzy operations such as the product t-norm operators, and that the Boolean values are replaced with fuzzy scores. In the neurosymbolic setting, those scores are determined by a neural network. This results in a computation that can be modelled using a single algebraic circuit.

\begin{example}{\sffamily Algebraic circuits: fuzzy propositional logic circuits}
 As example, consider the product t-norm operators, i.e., $x \oplus y = x + y - xy$ and $x \otimes y = xy$, and the formula \(\varphi_\mathbb{F} = a(V)_\mathbb{F} \land_\mathbb{F} (b(V)_\mathbb{F} \lor_\mathbb{F} c(V)_\mathbb{F})\).
 The following algebraic circuit computes the fuzzy score of $\varphi_\mathbb{F}$ for a given variable assignment to $V$, using neural networks to determine the fuzzy score.

    \begin{center}
        \begin{minipage}{.4\linewidth}
            \centering
            \begin{tikzpicture} [node distance=0.9cm]

    \node (n1) [plus] {$\otimes$};
    \node (na) [leaf, below of=n1, left of=n1, xshift=0.2cm] {$\lf_a$};
    \node (n2) [times, below of=n1, right of=n1, xshift=-0.2cm] {$\oplus$};

    \node (nna) [leaf, below of=n2, left of=n2, xshift=0.2cm] {$\lf_b$};
    \node (nb) [leaf, below of=n2, right of=n2, xshift=-0.2cm] {$\lf_{c}$};

    \draw [dirarrow] (na) -- (n1);
    \draw [dirarrow] (n2) -- (n1);
    \draw [dirarrow] (nna) -- (n2);
    \draw [dirarrow] (nb) -- (n2);
\end{tikzpicture}
  
        \end{minipage}
        \begin{minipage}{.5\linewidth}
            \begin{minipage}{.2\linewidth}
                where
            \end{minipage}
            \begin{minipage}{.3\linewidth}
                \begin{align*}
                    \lf_{a(V)}(V) &= nn_a(V) \\
                    \lf_{b(V)}(V) &= nn_b(V) \\
                    \lf_{c(V)}(V) &= nn_c(V)
                \end{align*}
            \end{minipage}
        \end{minipage}
    \end{center}
\end{example}

While such an algebraic circuit is similar in structure to Figure~\ref{fig:circuit-exa}, the semantics are very different. Indeed, the properties that were present in the probabilistic logic case, and that were exploited to transform Equation~\ref{eq:functional_dpl_example} into the algebraic circuit in Figure~\ref{fig:circuit-exa} are no longer present in the fuzzy setting.
In this regard, one can view the approach as a probabilistic fuzzy logic setting where the probability mass is assumed to be concentrated on one specific fuzzy assignment; an assignment that is selected by the neural networks. As exactly one assignment has a non-zero density, the aggregation (in this case an integration over the possible fuzzy values) reduces to a single assignment. For a more detailed explanation on this, we refer to Appendix~\ref{app:fuzzy_rewrite}.

In summary, both neurosymbolic probabilistic logic and fuzzy logic can be modelled using a single algebraic circuit. Even though the computation is similar, we do stress the difference in semantics. 
In probabilistic logic, neural networks parametrise the probability of assignments while in traditional fuzzy logic they output only a single fuzzy assignment.


\subsection{Efficient Implementation of Algebraic Circuits.}
In addition to research on theoretical properties of algebraic circuits, there has also been research on  efficient evaluation, and on the acceleration on modern hardware of these circuits. To name a few, \citet{dang2021juice} and \citet{klay} map a circuit evaluation to a set of layer evaluations; \citet{liuscaling} focus on probabilistic circuits specifically, which use for $\oplus$ a weighted sum, while \citet{shah2020acceleration,shah2021dpu} focus on developing specialized hardware accelerators. 

The DeepLog implementation integrates several tools to construct algebraic circuits. In particular, inference in probabilistic logic benefits from Boolean circuits that satisfy the d-DNNF properties. To obtain such circuits, knowledge compilation tools can be used. DeepLog currently integrates a compiler targeting sentential decision diagrams~\citep{pysdd}, and we plan to extend support for easily adding Decision-DNNF compilers~\citep{Lagniez17D4}. For the execution of algebraic circuits we use KLay~\citep{klay}, which enables efficient repeated evaluation on GPU.

The optimization of formulae into more efficient algebraic circuits is a broad, challenging task. 
While DeepLog currently does provide integration with knowledge compilation toolboxes and with KLay, automatically detecting without instructions that these are applicable on a given formulae is left for future work.
To facilitate the automation of this process in the future, we plan to define within a DeepLog model the properties that can be used to perform optimizations. For example, the semiring properties (commutativity, distributivity, associativity, neutral and annihilating elements) of algebraic structure \(\mathbb{P}\), that are required to optimize the neurosymbolic probabilistic logic setting into a single, efficient algebraic circuit.

\section{Evaluation}

The key contribution of  DeepLog is that it is 
 a neurosymbolic machine that is able to express
 and execute  a wide range of existing neurosymbolic AI systems that combine logic and neural networks through the DeepLog language and its algebraic circuits and software.
 It is declarative in that it is possible to emulate
 different neurosymbolic AI approaches by changing the algebraic structures and labelling functions.

 This section provides evidence for these claims.  More 
 specifically, we demonstrate the generality and flexibility of DeepLog by comparing some common approaches to neurosymbolic AI, as identified in Section \ref{sec:analysation}. We focus on two dimensions: 1) comparing logic in the architecture to logic in the loss, and 2)  comparing probabilistic, fuzzy, and probabilistic-fuzzy logics under different semantics. 
In DeepLog, the differences between such systems (e.g. DeepProbLog \citep{manhaeve_neural_2021} with logic in the architecture versus Semantic Loss \citep{xu2018semantic} with logic in the loss) are reduced to only a few lines of code, with all other factors held constant. 
{Furthermore, we look into the speed-up gained by the architectural choices made in the DeepLog implementation.}

While reading through the experimental section, the reader should keep in mind that although the experimental results are interesting by themselves w.r.t. the two dimensions mentioned above,  they are not intended as the definite answers to these questions. The experimental section is mainly meant
as a demonstration of the abilities of the DeepLog neurosymbolic machine. 
The other point to keep in mind is that while the current version of the software provides a first 
implementation, 
 there is ample room for further optimizing the circuits, for using approximate rather than exact inference on the circuits, and for sampling based inference based on the semantics provided through the algebraic circuits. 


\paragraph{Experimental details.}All experiments use algebraic circuits implemented in KLay \citep{klay}. For all metrics, we report the median, first and third quantiles over 10 repetitions, and measure statistical significance using the one-tailed Wilcoxon Signed-Rank Test with a significance level of 0.05. We train with the negative log-likelihood loss using the AdamW optimizer \citep{loshchilov2017adamW} and perform early stopping on the validation loss. For simplicity, we do not weigh the different losses when considering logic in the loss.
Further experimental details are provided in Appendix \ref{app:experiments}.

\subsection{Tasks}
We compare the methods on the following two tasks.

\paragraph{Visual Sudoku.}Visual Sudoku is a common neurosymbolic task where the aim is to predict the validity of a Sudoku, given the digits as images. A Sudoku is valid if it satisfies the three sets of constraints that ensure digits on the same row, column, and in the same box are all different.
We showcase the average precision and training time on 4x4 visual sudokus in Table~\ref{tab:sudoku}.

\paragraph{MNIST Addition.}MNIST Addition is a well-known neurosymbolic benchmark. Given two sequences of $N$ MNIST digits, the task is to predict the sum (e.g. $\digit{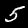}\digit{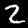}$ + $\digit{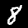}\digit{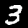}$ = 135).
This is challenging because the space of possible sums is combinatorial in the length $N$. Hence, the reasoning difficulty of this task can easily be increased by incrementing the number of digits in each sequence. Table \ref{tab:mnist-accuracy} shows the accuracy on this task. 

Recently, an alternative encoding of the MNIST Addition task was adopted~\citep{vankrieken2022anesi, maene2024deepsoftlog}. This encoding performs carry-based addition, leveraging repeating structure in the logic. Listings \ref{lst:mnist-original} and \ref{lst:mnist-carry} in Appendix \ref{app:experiments} show the original and carry-based programs, respectively.  The carry-based circuit is significantly smaller and, therefore, not directly comparable to the original. Note that this type of reduction in problem complexity is not possible in general. Table~\ref{tab:mnist-carry} shows the accuracy and inference time of the two best-performing approaches on the carry-based program.

\begin{table}[H]
    \centering
    \begin{tabular}{l r r}
        \toprule
        \multicolumn{1}{c}{} & Average precision (\%) & Training time (s) \\
         \toprule
         \textbf{Logic in the architecture} & & \\
         Probabilistic & $\textbf{99.71}_{-0.03}^{+0.04}$ & $\phantom{0}54.20_{-9.43}^{+8.64}$\\
         Fuzzy &  &\\
         \textit{\quad Gödel} &  $99.23_{-0.08}^{+0.41}$& $\phantom{0}44.35_{-2.78}^{+4.03}$\\
         \textit{\quad Lukasiewicz} & $50.00_{-0.00}^{+0.00}$ & $\phantom{00}9.15_{-0.25}^{0.08}$\\
         \textit{\quad Product} & $\textbf{99.69}_{-0.13}^{+0.03}$ & $114.61_{-31.69}^{+12.22}$\\
         Probabilistic-fuzzy &  &\\
         \textit{\quad Gödel} &  $50.29_{-1.56}^{+2.41}$ & $\phantom{0}72.22_{-5.33}^{+7.67}$\\
         \textit{\quad Lukasiewicz} & $50.00_{-0.00}^{+0.00}$ & $\phantom{0}12.36_{-0.09}^{+0.06}$\\
         \textit{\quad Product} & $\textit{99.02}_{-48.01}^{+0.18}$ & $\textit{132.43}_{-114.27}^{+5.57}$\\
         \midrule
         \textbf{Logic in the loss} & &\\
         Probabilistic & $52.70_{-2.12}^{+1.42}$ & $\phantom{00}6.12_{-1.88}^{+5.30}$\\
         Fuzzy & &\\
         \textit{\quad Gödel} & $\textbf{58.30}_{-4.01}^{+0.62}$ & $\phantom{0}40.82_{-1.67}^{+3.71}$\\
         \textit{\quad Lukasiewicz} & $51.74_{-0.20}^{+1.16}$ & $\phantom{00}9.78_{-5.06}^{+6.70}$\\
         \textit{\quad Product} & $52.28_{-1.46}^{+1.53}$ & $\phantom{0}13.00_{-5.18}^{+3.13}$\\
          Probabilistic-fuzzy & &\\
         \textit{\quad Gödel} & $50.65_{-0.84}^{+0.90}$ & $\phantom{0}21.53_{-7.80}^{+11.93}$\\
         \textit{\quad Lukasiewicz} & $51.58_{-1.16}^{+1.51}$ & $\phantom{0}13.53_{-7.24}^{+2.41}$\\
         \textit{\quad Product} & $51.73_{-0.83}^{+0.67}$& $\phantom{0}21.07_{-15.29}^{+10.66}$\\
         \midrule
         \textbf{Neural baseline} & $52.71_{-1.87}^{+1.63}$ & $\phantom{00}1.82_{-0.21}^{+0.63}$\\
         \bottomrule
    \end{tabular}
    \caption{In terms of average precision on the Visual Sudoku 4x4 test set, approaches that use logic in the architecture outperform logic in the loss methods. Probabilistic logic in the architecture trained faster than the product t-norm. Both train slower than the neural baseline, but perform significantly better. Boldface indicates significantly better results under the Wilcoxon Signed-Rank Test. The probabilistic-fuzzy product t-norm for logic in the architecture sometimes fails to find the global minimum, indicated with italics.}
    \label{tab:sudoku}
\end{table}


\begin{table}[H]
    \centering
    \begin{tabular}{l r r r r}
        \toprule
        \multicolumn{1}{c}{\textbf{Accuracy [$\%$]}} & \multicolumn{4}{c}{Number of digits} \\
        \cmidrule(lr){2-5}
         & 1 & 2 & 3 & 4\\
         \toprule
         \textbf{Logic in the architecture} & & &  \\
         Probabilistic              & $\textbf{97.90}_{-0.27}^{+0.12}$ & $95.64_{-0.20}^{+0.31}$            & $\textbf{94.21}_{-0.51}^{+0.08}$   & $\textbf{91.76}_{-0.18}^{+0.32}$  \\
         Fuzzy &  & &  \\
         \textit{\quad Gödel}       & $64.18_{-10.42}^{+1.98}$          & $36.54_{-13.25}^{+5.44}$          & $4.74_{-1.71}^{+8.34}$    & $0.32_{-0.30}^{+0.44}$  \\
         \textit{\quad Lukasiewicz} & $0.90_{-0.00}^{+0.00}$            & $0.00_{-0.00}^{+0.00}$            & $0.00_{-0.00}^{+0.00}$    &  $0.00_{-0.00}^{+0.00}$   \\
         \textit{\quad Product}   & $\textbf{97.87}_{-0.17}^{+0.12}$ & $\textbf{96.08}_{-0.17}^{+0.25}$   & $\textbf{94.06}_{-0.45}^{+0.06}$   & $\textbf{91.72}_{-0.24}^{+0.50}$  \\
         Probabilistic-fuzzy & & & \\
         \textit{\quad Gödel}       & $58.43_{-5.46}^{+6.59}$          & $1.78_{-0.41}^{+2.30}$             & $0.09_{-0.08}^{+0.03}$    & $0.00_{-0.00}^{+0.00}$ \\
         \textit{\quad Lukasiewicz} & $7.72_{-0.73}^{+0.22}$           & $0.00_{-0.00}^{+0.00}$             & $0.00_{-0.00}^{+0.00}$    & $0.00_{-0.00}^{+0.00}$  \\
         \textit{\quad Product}   & $\textbf{97.74}_{-0.14}^{+0.07}$ & $95.32_{-0.24}^{+0.35}$   & $93.16_{-1.02}^{+0.11}$   & $89.68_{-1.12}^{+0.16}$\\
         \midrule
         \textbf{Logic in the loss} & & &  \\
         Probabilistic              & $97.26_{-0.15}^{+0.27}$           & $83.66_{-1.17}^{+2.19}$            & $1.86_{-0.80}^{+0.63}$  & $0.04_{-0.04}^{+0.46}$  \\
         Fuzzy & & &  \\
         \textit{\quad Gödel}       & $97.52_{-0.18}^{+0.25}$           & $87.16_{-2.19}^{+1.50}$            & $3.03_{-1.76}^{+0.84}$  & $0.00_{-0.00}^{+0.06}$  \\
         \textit{\quad Lukasiewicz} & $37.27_{-21.91}^{+55.81}$        & $83.06_{-2.83}^{+1.91}$                   & $1.11_{-0.72}^{+0.08}$  & $0.00_{-0.00}^{+0.00}$  \\
         \textit{\quad Product}   & $97.38_{-0.14}^{+0.14}$           & $88.42_{-1.37}^{+1.25}$            & $1.29_{-0.21}^{+0.51}$  & $0.24_{-0.16}^{+0.60}$ \\
          Probabilistic-fuzzy & & & \\
         \textit{\quad Gödel}       & $97.42_{-0.20}^{+0.09}$           & $74.64_{-3.39}^{+1.98}$                     & $0.06_{-0.00}^{+0.89}$  & $0.00_{-0.00}^{+0.00}$  \\
         \textit{\quad Lukasiewicz} & $10.24_{-0.00}^{+0.00}$          & $82.08_{-1.06}^{+3.17}$                     & $1.17_{-0.99}^{+0.08}$  & $0.00_{-0.00}^{+0.00}$  \\
         \textit{\quad Product}   & $97.19_{-0.24}^{+0.36}$          & $86.36_{-1.08}^{+1.50}$            & $3.09_{-1.68}^{+1.65}$  & $1.00_{-0.34}^{+0.82}$  \\
         \midrule
         \textbf{Neural baseline}   & $97.46_{-0.28}^{+0.18}$          & $79.36_{-1.45}^{+2.88}$                     & $1.02_{-0.83}^{+0.18}$  & $0.00_{-0.00}^{+0.00}$ \\
         \bottomrule
    \end{tabular}
    \caption{The highest accuracy on the MNIST Addition test set, indicated with boldface, is achieved by the probabilistic semiring and product t-norm used in the architecture. Using logic in the loss only slightly improves the neural baseline.}
    \label{tab:mnist-accuracy}
\end{table}

\renewcommand{\thefootnote}{\fnsymbol{footnote}}
\begin{table}[H]
    \centering
    \begin{tabular}{l r r r r}
        \toprule
        \multicolumn{1}{c}{} & \multicolumn{4}{c}{Number of digits} \\
        \cmidrule(lr){2-5}
         & 1 & 2 & 4 & 10\\
         \toprule
         \textbf{Accuracy [$\%$]} & & & \\
         Probabilistic  & $\textbf{97.71}_{-0.13}^{+0.41}$ & $\textbf{95.56}_{-0.16}^{+0.20}$ & $\textbf{92.00}_{-0.48}^{+0.46}$ & $\textbf{85.00}_{-0.55}^{+1.10}$ \\
         Product       & $\textbf{97.83}_{-0.21}^{+0.12}$ & $\textbf{95.82}_{-0.26}^{+0.41}$ & \textit{$\textbf{92.20}_{-1.08}^{+0.74}$}\tablefootnote{Two out of ten runs failed to converge.   \label{fn:convergence}} & \textit{$\textbf{85.20}_{-0.55}^{+0.85}$}\textsuperscript{\ref{fn:convergence}} \\
         Neural baseline & $97.36_{-0.18}^{+0.15}$ & $83.66_{-3.36}^{+3.04}$ & $0.00_{-0.00}^{+0.08}$ & $0.00_{-0.00}^{+0.00}$ \\
         \midrule
         \textbf{Inference time [ms/batch]} & & &  \\
         Probabilistic   & $5.15_{-0.06}^{+0.16}$   & $5.21_{-0.06}^{+0.11}$  & $5.40_{-0.01}^{+0.09}$  & $6.06_{-0.09}^{+0.18}$  \\
         Product-T & $13.49_{-0.11}^{+0.10}$   & $13.72_{-0.14}^{+0.13}$   & $14.23_{-0.15}^{+0.16}$  & $15.83_{-0.19}^{+0.23}$  \\
         Neural baseline & $0.38_{-0.01}^{+0.01}$    & $0.35_{-0.01}^{+0.01}$  & $0.36_{-0.00}^{+0.02}$  & $0.43_{-0.00}^{+0.00}$ \\
         \bottomrule
    \end{tabular}
    \caption{The accuracy on the MNIST Addition test set using the carry-based encoding and logic in the architecture. The product t-norm sometimes fails to find the global minimum for larger problem sizes, indicated with italics. Boldface indicates significantly better results under the Wilcoxon Signed-Rank Test.}
    \label{tab:mnist-carry}
\end{table}
\renewcommand{\thefootnote}{\arabic{footnote}}

Usually, NeSy systems are directly compared, although they might differ in more than one aspect and have different implementations. DeepLog’s unifying theoretical framework allows each aspect of a NeSy system to be compared in isolation using a common implementation, making it possible to easily answer fundamental questions such as the following.\\

\noindent\textbf{RQ 1. Which differences in task performance and computational cost can be attributed to logic in the architecture compared to logic in the loss?}

Across all tasks, logic in the architecture achieves the best performance, while logic in the loss only slightly improves the neural baseline. Logic in the loss guides optimization towards better satisfaction of the constraints, but still requires the neural network to solve the entire task. In contrast, logic in the architecture provides hard guarantees of rule satisfaction conditioned on the neural network's predictions, providing a strong inductive bias. In this case, the logic solves part of the problem (e.g. enforcing Sudoku rules and performing addition) while the neural network only has to solve the remaining perception task. As the MNIST-Addition task (Table \ref{tab:mnist-accuracy} and \ref{tab:mnist-carry}) becomes harder, the neural network 
has increasingly more tasks to solve and less supervisions for each of them, leading to a drop in performance of the baseline and the logic in the loss approach. The performance of logic in the architecture remains high, as the neural network is still only required to solve the perception task, which did not get harder. Any performance loss can, therefore, be attributed to compounding  errors on the neural network's predictions. Finally, we note that, as logic in the loss only uses the circuit during training, the inference time is equal to that of the neural baseline. The performance of logic in the architecture comes at a cost, as logic and probabilistic inference are still executed for each query after the neural network has made its predictions. \\

\noindent\textbf{RQ 2. Which differences in task performance and computational cost can be attributed to the choice of semantics?}

Probabilistic semantics consistently achieve the best task performance. Under fuzzy semantics, both the Gödel and Łukasiewicz T-norm suffer from the \textit{vanishing derivative problem} previously identified by \cite{van2022analyzing}. This problem implies that the derivative of some of these operators becomes zero within some interval, thereby losing the learning signal. Using logic in the loss largely circumvents the vanishing derivative problem, as there is always a learning signal due to the neural network's direct predictions. The product t-norm does not have a vanishing derivative and generally achieves the best performance under fuzzy semantics. Its performance is often comparable to probabilistic semantics, as their underlying computation is similar for most of our tasks. However, unlike in the probabilistic setting, the product t-norm assumes all proofs are independent,  which leads to a systematic difference with the probabilistic setting. %
On the MNIST Addition task with carry-based encoding (Table \ref{tab:mnist-carry}), this causes the product t-norm to fail to find the global minimum on two out of ten runs for $N=4$ and $N=10$. The probabilistic-fuzzy approach typically performs slightly worse than fuzzy. On the Sudoku task (Table \ref{tab:sudoku}), probabilistic-fuzzy fails to optimize, because the models tend to get stuck in a local minimum where the probability distribution converges to always predict equal fuzzy scores for the digit classes, regardless of the input image. Sudokus contain symmetries, so the classifier can learn any permutation of symbols to image classes, which can result in not learning any specific matching in the case of probabilistic fuzzy. 

Computationally, fuzzy semantics are generally less expensive than probabilistic reasoning, as they do not require circuit compilation to be sound. Smaller circuit sizes resulting from this often make fuzzy circuits faster to construct and train with. However, in both encodings of the MNIST addition task, all proofs are mutually exclusive, eliminating the need for compilation and making the probabilistic and fuzzy circuits the same size. The product t-conorm operator requires multiple operations (i.e. $a + b - a*b$) per node instead of one (i.e. $a+b$), explaining the difference in timings. \\

{\noindent\textbf{RQ 3. How does the performance of DeepLog’s reimplementation of DeepProbLog compare to the original implementation?}}

{
Currently, the uptake  of NeSy systems is hindered by the high computational cost of inference in contemporary neurosymbolic systems. However, this is often an implementation issue that can be avoided by making the right architectural choices.
For our DeepLog implementation, we compile everything into PyTorch modules, avoiding any bookkeeping by replacing them with appropriate gather and scatter operations. The final circuit should be free of any CPU-bound logic. To show the effectiveness of this approach,
we compare the original implementation of DeepProbLog~\citep{manhaeve2018deepproblog, manhaeve_neural_2021} to a reimplementation in DeepLog for the MNIST Addition task on both CPU and GPU.
The results are shown in Table \ref{tab:inference-time-dpl}. Although knowledge compilation can be  time-intensive, it only has to be performed once, and is therefore excluded from the time reported here. For DeepLog, we use a batch size that achieves optimal resource utilization. 
For the original DeepProbLog implementation, we only show the result for the CPU as evaluating the neural network on the GPU is slower. This is due to the fact that the original implementation does not batch, but rather handles each example sequentially. Any speed-up of evaluating the neural network on the GPU is lost due to the overhead of switching between the CPU and the GPU.
Note that DeepLog on the CPU is already several orders of magnitude faster than the original DeepProbLog implementation. Because DeepLog compiles   to a plain PyTorch module, using a GPU further increases the speed of inference.
Note that this not only due to the use of KLay. Although KLay offers a great speed-up in the evaluation of the circuit, the entire computational graph needs to be well implemented for the final computation to be efficient. 
}

\begin{table}[H]
  \centering
  
  \begin{tabular}{@{}c c c c@{}}
    \toprule
    \multicolumn{1}{c}{} & \multicolumn{3}{c}{Inference time [s/query]} \\
    \cmidrule(lr){2-4}
    Digits &
    DeepProbLog CPU &
    DeepLog CPU &
    DeepLog GPU \\
    \midrule
    1 & $6.14\times10^{-3}\,\pm\,5.85\times10^{-4}$
      & $1.09\times10^{-5}\,\pm\,6.46\times10^{-7}$
      & $5.62\times10^{-7}\,\pm\,9.10\times10^{-8}$ \\
    2 & $3.52\times10^{-2}\,\pm\,1.28\times10^{-2}$
      & $4.13\times10^{-5}\,\pm\,3.72\times10^{-6}$
      & $1.52\times10^{-6}\,\pm\,1.37\times10^{-7}$ \\
    3 & $1.32\times10^{-1}\,\pm\,4.74\times10^{-2}$
      & $2.43\times10^{-3}\,\pm\,2.36\times10^{-4}$
      & $1.52\times10^{-4}\,\pm\,2.07\times10^{-5}$ \\
    4 & \multicolumn{1}{c}{--}
      & $1.96\times10^{-1}\,\pm\,7.80\times10^{-3}$
      & $5.40\times10^{-2}\,\pm\,1.02\times10^{-4}$ \\
    \bottomrule
  \end{tabular}
  \caption{The inference time in seconds per query ($\frac{batch\_time}{batch\_size}$), which includes the evaluation of both the neural network and the algebraic circuit. The average and standard deviation are provided, over 100 batches. For each entry, the batch size is chosen for optimal resource allocation, as shown in Table~\ref{tab:optimal-batch}. The most performant solution is to use DeepLog with GPU acceleration.}
  \label{tab:inference-time-dpl}
\end{table}

\section{Related Work}



Neurosymbolic AI focuses on enhancing neural models with reasoning capabilities and enriching symbolic models with perceptual components.
The degree and nature of integrated neural - symbolic models varies widely—from loosely coupled, hand-engineered pipelines to tightly integrated probabilistic models.
The diversity of both the neural and the symbolic components has resulted in a wide range of different neurosymbolic AI system -- an alphabet soup of systems \citep{ serafini_logic_2016, diligenti2017sbr, sourek2018lrnn, manhaeve2018deepproblog, xu2018semantic, si2019difflog, marra2019integrating, yang2020neurasp,  ahmed2022semantic}.  The focus on systems has also lead to an increasing emphasis
on empirical evaluation, at the cost of a deeper theoretical understanding of  neurosymbolic AI.   Nevertheless, there is a growing awareness in the community 
about these issues and several interesting results that we will now discuss. 

While discussing these results, and contrasting them with the DeepLog neurosymbolic machine, it is important to keep in mind the six distinguishing features of DeepLog mentioned at the end of the introduction. The most important differences between DeepLog 
and other approaches are that DeepLog is not only descriptive, but it is a neurosymbolic machine that consists of a mathematical and computational framework that is not only operational but also allows to emulate a wide range of neurosymbolic approaches.

First, 
there are several attempts to  order the existing landscape of systems in a systematic manner.  In Section \ref{sec:analysation}, we  already introduced the categorization of \cite{marra2024statistical}, where neurosymbolic systems are categorized according to 7 dimensions inherited from the statistical relational AI field. The framework has proven to be highly effective in capturing different nuances of the current neurosymbolic landscape and helps to establish meaningful connections between systems that superficially seem distinct.
Another noteworthy attempt is the one from \cite{van_bekkum_modular_2021}, where a boxology language is used to describe NeSy approaches and systems as different compositions of standard boxes encapsulating neurosymbolic learning and reasoning components. These boxes are like design patterns and the boxology framework focusses on modularity. Both frameworks are descriptive rather than formal, mathematical let alone operational.  While they are able to provide high-level descriptions and abstractions of multiple systems using common blocks, dimensions or functions, they cannot be used to design and construct new systems.

Second, 
instead of categorising neurosymbolic systems, other efforts aim for a unification of the field in various directions.
There exists a long tradition in studying how neural networks can encode logic
and vice versa, how logic can be used to encode neural networks and vice versa
how neural networks can encode various types of logic \citep{avila1999connectionist,holldobler1999approximating}. In this tradition, 
\citet{odense2025semantic} contribute a semantic framework and theory that determines when and if neural networks encode a logical semantics, which can be used to translate a logic into a neural network. While logics and semantics are part of \dl,  they are directly operationalised as algebraic circuits at a logical level, not as neural networks. Furthermore, logic is used as a separate layer to enforce logical consistency on top of neural network outputs instead of obtaining a semantically equivalent neural network with symbolic inputs and outputs. While the work of \citet{odense2025semantic} is at the semantic level only, \dl also offers an operational and computational framework.


Third,
the framework closest in spirit to \dl are the neurosymbolic energy-based models (\nesyebms) of~\citet{dickens2024mathematical}, yet they are not as operational as \dl.
\nesyebms posit abstract energy functions as the central quantity of neurosymbolic models and inference.
That is, each neurosymbolic system defines its own energy function, and so an energy-based model, determined by the internals of the system. 
This unified view of neurosymbolic systems as \nesyebms enables different systems to share the same algorithms for inference and learning. However, unlike \dl that integrates systems by exposing and recombining their fundamental semantic components (i.e. labelling functions), \nesyebms do not decompose systems; they collapse each system into an  energy term, obscuring the internal structure and limiting operational transparency.
In contrast, \dl further generalises the definitions of~\citet{de2025defining} and makes them \emph{fully} operational from input to output;
given a sentence, a choice of semantics, and an inference task from any neurosymbolic model, \dl produces a uniform, algebraic representation that directly implements the desired label as an algebraic circuit.

Finally,  
ULLER~\citep{Uller24} aims to be a unifying \qq{lingua franca} that can be interpreted by different neurosymbolic systems, which all employ their own specific formalism.
ULLER acts as an interface and allows for knowledge and data to be encoded in one common format and used by many systems, i.e the result is a uniform interface and representation with non-uniform computation. It is neither operational nor is it a neurosymbolic machine that can be used for reasoning about and optimizing inference and learning.
ULLER does not  define a single  semantics, but rather leaves the semantics as an open choice determined by the NeSy system that is interfaced.
\dl is the opposite; it allows the user to encode knowledge in their language of choice and compiles it down into the same intermediate \dl language that operationalises as algebraic circuits.
That is, \dl is a neurosymbolic machine, with a uniform representation and computational mechanism.




\section{Conclusion}
We have introduced DeepLog, a neurosymbolic machine that captures a diverse set of approaches within a single theoretical and operational framework. At its core, DeepLog is built around a common intermediate language, which abstracts over specific logics—Boolean, fuzzy, or probabilistic—and over implementation choices such as whether logic appears in the architecture or in the loss. This intermediate representation allows systems to be expressed in a uniform way, making their semantics  explicit and  comparable.

 DeepLog operationalizes the intermediate level through algebraic circuits. By compiling the intermediate representations into modular, composable computational graphs, DeepLog provides a principled basis for controlled comparisons between different neurosymbolic systems. This modular design not only clarifies differences and commonalities, but also facilitates reproducibility and systematic exploration of design trade-offs.

Looking ahead, we see several promising directions. We plan to develop automatic circuit optimizations and extend the DeepLog implementation to support approximate inference. We also envision DeepLog as a foundation for higher-level languages and libraries, enabling broader adoption and fostering an interoperable ecosystem. By providing both a unifying theory and a practical platform, DeepLog aims to enable the field of neurosymbolic AI to move from a fragmented collection of systems into a more coherent and principled discipline.

\emph{The DeepLog Software will be released at a later point in time, in any case after the paper has been submitted and accepted, if not before. Please contact the authors.}

\section*{Acknowledgements}
This work project has received funding from the European Research Council (ERC) under the European Union’s Horizon 2020 research and innovation programme (Grant agreement No. 101142702).
This research received funding from the Flemish Government under the “Onderzoeksprogramma Artificiële Intelligentie (AI) Vlaanderen” programme. 
The authors thank Stefano Colamonaco, Jaron Maene and Pedro Zuidberg Dos Martires for their input and feedback.

\clearpage
\appendix

\section{Modelling Dependencies}
\label{app:dependencies}

As discussed in Section~\ref{sec:composing_ac}, several neurosymbolic systems assume a factorized weight function to improve inference tractability. For example, in 
\begin{align*}
\varphi_\mathbb{P}(B, E, V, S) = 
    \sum_{B} \sum_{E} (b(V)[B]_\mathbb{B} \lor_\mathbb{B}
        e(S)[E]_\mathbb{B})_\mathbb{P} \times_\mathbb{P}
        (b(V)[B]_\mathbb{P} \times_\mathbb{P} e(S)[E]_\mathbb{P})
\end{align*}
we have \(b(V)[B]_\mathbb{P} \times_\mathbb{P} e(S)[E]_\mathbb{P}\). This assumption is without loss of generality due to the presence of the logical component, here \(b(V)[B]_\mathbb{B} \lor_\mathbb{B} e(S)[E]_\mathbb{B}\), 
and our ability to add more atoms. We will explain how to model a dependency even with a factorized weight function using an example. For more information, we refer to \citet{ChaviraD08}, who show how to model Bayesian networks using such an approach, and to Example 13 of \citet{Derkinderen24Semiring}.

Consider the case where the occurrence of a burglary does depend on the occurrence of an earthquake written in DeepProbLog syntax:
\begin{lstlisting}[mathescape=true, captionpos=b, caption={MNIST Addition original encoding.}, frame = single]
$\mathsf{nn(classifier1, Video) :: burglary(Video).}$
$\mathsf{nn(classifier2a, Seismic) :: e\_if\_b(Seismic).}$
$\mathsf{nn(classifier2b, Seismic) :: e\_if\_not\_b(Seismic).}$

$\mathsf{earthquake(Seismic) \colonminus 
    (e\_if\_b(Seismic), burglary(Video));( e\_if\_not\_b(Seismic), not(burglary(Video)))}$.

$\mathsf{?- burglary(v1); earthquake(s1).}$
\end{lstlisting}
The relation between \(earthquake\), \(e\_if\_b\), and \(e\_if\_not\_b\) is encoded using logic:
\begin{align*}
\varphi_\mathbb{P}(V, S) = 
    \sum_{B} &\sum_{E} \sum_{E'} \sum_{E''} 
        \Big( \\
            &\quad (b(V)[B]_\mathbb{B} \lor_\mathbb{B} e(S)[E]_\mathbb{B}) \land \\
            &\quad (e(S)[E]_\mathbb{B} \iff_\mathbb{B} (b(V)[B]_\mathbb{B} \land_\mathbb{B} e\_if\_b(S)[E']_\mathbb{B} \lor_\mathbb{B} \neg_\mathbb{B} b(V)[B]_\mathbb{B} \land_\mathbb{B} e\_if\_not\_b(S)[E'']_\mathbb{B})) \\
        &\Big)_\mathbb{P} \times_\mathbb{P} (b(V)[B]_\mathbb{P} \times_\mathbb{P} e(S)[E]_\mathbb{P} \times_\mathbb{P} e\_if\_b(S)[E']_\mathbb{P} \times_\mathbb{P} e\_if\_not\_b(S)[E'']_\mathbb{P}),
\end{align*}
where we use \(\iff_\mathbb{B}\) as a shorthand for convenience. To obtain the correct probability, the weight function in this case is encoded here as

\begin{align*}
    \lf(b(V)[B]_\mathbb{P}) &:= 
    \begin{cases}
        nn_1(V) \ \text{ if } B \qequal true, \\
        1-nn_{1}(V) \ \text{ otherwise,}
    \end{cases} \\
    \lf(e(S)[E]_\mathbb{P};I) &:= 1.0 \\
    \lf(e\_if\_b(S)[E']_\mathbb{P};I) &:= 
    \begin{cases}
        nn_{2'}(S) \ \text{ if } E' \qequal true, \\
        1-nn_{2'}(S) \ \text{ otherwise,}
    \end{cases} \\
    \lf(e\_if\_not\_b(S)[E'']_\mathbb{P}) &:= 
    \begin{cases}
        nn_{2''}(S) \ \text{ if } E'' \qequal true, \\
        1-nn_{2''}(S) \ \text{ otherwise,}
    \end{cases} \\
\end{align*}
The logic labelling function acts as a simple identity function; for any reified algebraic atom \(x[Y]_\mathbb{B}\):
\begin{align*}
    \lf(x[Y]_\mathbb{B}) &:= 
    \begin{cases}
        true \ \text{ if } Y \qequal true, \\
        false \ \text{ otherwise.}
    \end{cases} \\
\end{align*}

\section{Algebraic Circuits in Neurosymbolic Fuzzy Logic}
\label{app:fuzzy_rewrite}
A standard task in the neurosymbolic fuzzy logic setting is to evaluate \(\varphi_\mathbb{F}\), which is a formula that comprises fuzzy operations and where the interpretation of each atom is a fuzzy score that is determined by a neural network. This formula can trivially be mapped to a single evaluation of the algebraic circuit.

\begin{example}{}
 As example, consider the product t-norm operators, i.e., $x \oplus y = x + y - xy$ and $x \otimes y = xy$, and the formula \(\varphi_\mathbb{F} = a(V)_\mathbb{F} \land_\mathbb{F} (b(V)_\mathbb{F} \lor_\mathbb{F} c(V)_\mathbb{F})\).
 The following algebraic circuit computes the fuzzy score of $\varphi_\mathbb{F}$ for a given variable assignment, using neural networks to determine the fuzzy score.

    \begin{center}
        \begin{minipage}{.4\linewidth}
            \centering
            \begin{tikzpicture} [node distance=0.9cm]

    \node (n1) [plus] {$\otimes$};
    \node (na) [leaf, below of=n1, left of=n1, xshift=0.2cm] {$\lf_a$};
    \node (n2) [times, below of=n1, right of=n1, xshift=-0.2cm] {$\oplus$};

    \node (nna) [leaf, below of=n2, left of=n2, xshift=0.2cm] {$\lf_b$};
    \node (nb) [leaf, below of=n2, right of=n2, xshift=-0.2cm] {$\lf_{c}$};

    \draw [dirarrow] (na) -- (n1);
    \draw [dirarrow] (n2) -- (n1);
    \draw [dirarrow] (nna) -- (n2);
    \draw [dirarrow] (nb) -- (n2);
\end{tikzpicture}
  
        \end{minipage}
        \begin{minipage}{.5\linewidth}
            \begin{minipage}{.2\linewidth}
                where
            \end{minipage}
            \begin{minipage}{.3\linewidth}
                \begin{align*}
                    \lf_{a}(V) &= nn_a(V) \\ 
                    \lf_{b}(V) &= nn_b(V) \\ 
                    \lf_{c}(V) &= nn_c(V) 
                \end{align*}
            \end{minipage}
        \end{minipage}
    \end{center}
\end{example}

This approach can be viewed in the probabilistic context---probabilistic fuzzy logic---if we assume that the probability mass is concentrated on one specific fuzzy assignment; an assignment that is selected by the neural networks.
In this view, the above approach is a rewrite that more easily allows the use of algebraic circuits.

We briefly explain this probabilistic perspective in more detail as it helps to better contrast the fuzzy logic setting with the probabilistic logic setting. According to this perspective, the neurosymbolic fuzzy logic formula can be redefined as
\begin{equation}\label{eq:fuzzy_as_ac:def}
    \mathcal{T}(V) =
    \int\limits_{A_1} \ldots \int\limits_{A_n}
        (\varphi_\mathbb{F})_\mathbb{R} \times_\mathbb{R} 
        (\varphi'_\mathbb{P})_\mathbb{R} 
    dA_1 \dots dA_n,
\end{equation}
with \(\mathbb{R}\) an algebraic structure of real values that represent expected fuzzy scores, \(\varphi_\mathbb{F}\) a formula that represents the fuzzy score and that comprises atoms \(a_1(V)[A_1]_\mathbb{F}, \dots, a_n(V)[A_n]_\mathbb{F}\) whose truth values are fuzzy scores, and \(\varphi'_\mathbb{P}\) a formula comprising atoms \(a_1(V)[A_1]_\mathbb{P}, \dots, a_n(V)[A_n]_\mathbb{P}\), representing the probability density of interpretation \(\{a_1(V) {:=} A_1, \dots, a_n(v) {:=} A_n\}\).
To efficiently evaluate this equation, neurosymbolic fuzzy logic relies on the aforementioned assumption: the probability distribution is assumed to concentrate the probability mass on one specific assignment, the one indicated by neural network outputs $nn_i(V)$. Mathematically this is defined as
\begin{equation*}
    \lf(\varphi'(V)_\mathbb{P} = \prod_{i = 1}^n \lf(a_i(V)[A_i]_\mathbb{P}) = \prod_{i = 1}^n\delta(A_i - nn_i(V)).
\end{equation*}
Because of this probability assumption, the only relevant assignment to \( (A_1, \dots, A_n) \) is \( ( nn_1(V), \dots, nn_n(V) )\). Hence, Equation~\ref{eq:fuzzy_as_ac:def} is more efficiently computed as
\begin{equation}
    \varphi_\mathbb{F}(\{A_1 \mapsto nn_1(V), \dots, A_n \mapsto nn_n(V)\}),
\end{equation}
which, as stated earlier, is what neurosymbolic fuzzy logic systems typically do.



\begin{example}{}
 As example, consider
    \begin{multline}\label{eq:ex:prob-fuzzy-to-fuzzy}
        \int\limits_{A} \int\limits_{B}  \int\limits_{C} 
                \Big[a(V)[A]_\mathbb{F} \land_\mathbb{F} (b(V)[B]_\mathbb{F} \lor_\mathbb{F} c(V)[C]_\mathbb{F})\Big]_\mathbb{R}
                    \times_\mathbb{R} 
                \Big[a(V)[A]_\mathbb{P} \land_\mathbb{P} b(V)[B]_\mathbb{P} \land_\mathbb{P} c(V)[C]_\mathbb{P}\Big]_\mathbb{R} \\ dA \; dB \; dC,
    \end{multline}
    where \(a(V)\), \(b(V)\), and \(c(V)\) are atoms with a fuzzy truth value, which is determined by the integration. Their labels in a fuzzy context are the fuzzy scores themselves.
    \begin{equation*} 
        \lf(a(V)[A]_\mathbb{F}) := A \quad
        \lf(b(V)[B]_\mathbb{F}) := B \quad
        \lf(c(V)[C]_\mathbb{F}) := C
    \end{equation*}
    The probabilistic labels are defined using the Dirac delta.
    \begin{align*}
        \lf(a(V)[A]_\mathbb{P}) &:= \delta(A - nn_a(V)) \\
        \lf(b(V)[B]_\mathbb{P}) &:= \delta(B - nn_b(V)) \\
        \lf(c(V)[C]_\mathbb{P}) &:= \delta(C - nn_c(V))
    \end{align*}
    Because of these Dirac deltas, the probability density is non-zero only when
    \begin{equation*}
        \{  A \mapsto nn_a(V), \quad  B \mapsto nn_b(V), \quad C \mapsto nn_c(V) \}.
    \end{equation*}
    In other words, under the above variable assignment, Equation~\ref{eq:ex:prob-fuzzy-to-fuzzy} results into 
    \begin{equation}
        nn_a(V) \otimes (nn_b(V) \oplus nn_c(V)).
    \end{equation}
\end{example}

In summary, a traditional neurosymbolic fuzzy logic system (Equation~\ref{eq:fuzzy_as_ac:def}) computes the expected fuzzy score while assuming a Dirac-delta distribution, which is efficiently computable as a single circuit evaluation\footnote{As example of a less traditional neurosymbolic fuzzy logic systems, consider NeuPSL~\citep{NeuPSLPryor23}. They do not assume a Dirac delta distribution, and as a consequence their evaluation no longer translates to a single circuit evaluation.}.

\section{Experimental Setup}
\label{app:experiments}


\begin{table}[h]
    \centering
    \begin{tabular}{lll}
    \toprule
         & Visual Sudoku & MNIST-Addition\\
    \midrule
    GPU           &  NVIDIA RTX 3080 Ti & NVIDIA L40S\\
    batch size    & 64 & 16\\
    learning rate & $1e^{-3}$ & $1e^{-3}$\\
    patience      & 5 & 5\\
    neural network & LeNet & LeNet\\
    \bottomrule
    \end{tabular}
    \caption{Experimental details.}
    \label{tab:experimental-details}
\end{table}

\paragraph{Implementation of design choices.}The logic in the loss approach uses the same network as logic in the architecture, but has two separate heads: one predicts inputs to the logic (e.g., MNIST classification), and the other predicts the answer to the full task (e.g., Sudoku validity or the sum of the numbers) using the horizontally joined activations of all images. 

We implement probabilistic-fuzzy semantics by using a neural network to predict the parameters of a probability distribution over the fuzzy labels of the atoms. Samples from this distribution are then propagated through a circuit to compute the corresponding fuzzy label of the formula. Aggregating these results across samples yields the expected fuzzy value of the formula.



\paragraph{MNIST Addition encodings} Listing \ref{lst:mnist-original} and \ref{lst:mnist-carry} show the original and the carry-based MNIST Addition encoding, respectively.

The first main difference is that the carry-based encoding predicts the probability of each digit in the sum separately, thereby lowering the number of ground atoms that need to be proven. For example, for MNIST Addition with 2-digit numbers, the carry-based encoding computes the probability of $2 + 10 + 10 = 22$ ground atoms, compared to $199$ for the original encoding. The second difference is structural: in the original encoding, the \lstinline{number} predicate is the main source of circuit size with a fan-out of ten at each level. In the circuit-based encoding, only the \lstinline{carry} predicate is defined recursively. This predicate only has a fan-out of two at each level, resulting in significantly smaller circuits. 

Note that such a reduction in problem complexity is not possible in general, but, in this specific case, it leads to circuits that are structurally incomparable. Therefore, these two encodings should be treated as two distinct tasks.

\begin{lstlisting}[mathescape=true, captionpos=b, caption={MNIST Addition original encoding.}, label=lst:mnist-original, frame = single]
$\mathsf{classifier(I,N) :: classify(I,N).}$
$\mathsf{digit(I,N) \colonminus between(0, 9, N), classify(I,N).}$

$\mathsf{number([~], 0).}$
$\mathsf{number([H \mid T], N) \colonminus digit(H, N1), number(T, N2), length(T, L), is(N, (10}$**$\mathsf{L)}$*$\mathsf{N1+N2).}$

$\mathsf{addition(L1, L2, S) \colonminus number(L1, N1), number(L2, N2), is(S,N1+N2).}$

$\mathsf{?- addition([img0, img1], [img2, img3], S)}$
\end{lstlisting}

Note that here, for familiarity, we use an encoding inspired by DeepProbLog. This serves as an example of a high-level language that would get compiled into the DeepLog intermediate representation. For more info, see Appendix~\ref{app:syntax}.
A possible translation to DeepLog for the query \(\mathtt{addition(\digit{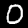},\digit{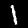},1)}\) could be:
\[
\varphi(I_1,I_2) = \sum_{D_1} \sum_{D_2} 
    \left(\mathtt{is}(D_1,D_2)_\mathds{B}\right)_\mathds{P}
    \times_\mathds{P} \left(\mathtt{classify}(I_1,D_1)_\mathds{P}
    \times_\mathds{P} \mathtt{classify}(I_2,D_2)_\mathds{P} \right)
\]






\begin{lstlisting}[mathescape=true, captionpos=b, caption={MNIST Addition carry-based encoding.}, label=lst:mnist-carry, frame = single]
$\mathsf{classifier(I,N) :: classify(I,N).}$
$\mathsf{digit(I,N) \colonminus between(0, 9, N), classify(I,N).}$

$\mathsf{addMod10(I1, I2, C, N) \colonminus between(0, 1, C), digit(I1, N1), digit(I2, N2), is(N, (N1+N2+C)~mod~10).}$
$\mathsf{addDiv10(I1, I2, Cin, Cout) \colonminus between(0, 1, Cin), digit(I1, N1), digit(I2, N2),  is(Cout, (N1+N2+Cin)~div~10).}$

$\mathsf{carry([~], [~], 0).}$
$\mathsf{carry([H1 \mid T1], [H2 \mid T2], Cout) \colonminus carry(T1, T2, Cin), addDiv10(H1, H2, Cin, Cout).}$

$\mathsf{addition([~], [~], 0).}$
$\mathsf{addition([H1 \mid T1], [H2 \mid T2], N) \colonminus carry(T1, T2, Cin), addMod10(H1, H2, Cin, N).}$

$\mathsf{?- carry([img0, img1], [img2, img3], N1)}$
$\mathsf{?- addition([img0, img1], [img2, img3], N2)}$
$\mathsf{?- addition([img1], [img3], N3)}$
\end{lstlisting}

\begin{table}[H]
    \centering
    \begin{tabular}{r r r r}
        \toprule
         \textbf{Digits} & DeepProbLog CPU & DeepLog CPU &  DeepLog GPU \\
         \midrule
          1 & 1 & 256 & 8192 \\
          2 & 1 & 128 & 1024 \\
          3 & 1 & 128 & \phantom{0}512 \\
          4 & 1 & \phantom{0}64 & \phantom{000}2 \\
         \bottomrule
    \end{tabular}
    \caption{Batch size that achieves optimal resource utilization on the MNIST Addition task.}
    \label{tab:optimal-batch}
\end{table}


\section{Syntax} \label{app:syntax}
In this section, we provide a concise overview of logic programming syntax.
A term~$t$ can be a constant~$c$, a variable~$V$, or a structured term of the form $f(u_1, \ldots, u_k)$, where $f$ is a functor and each $u_i$ is itself a term. 
Atoms are expressions of the form $q(t_1, \ldots, t_n)$, where $q$ is a predicate of arity $n$ (also denoted $q/n$) and each $t_i$ is a term.
A literal is either an atom or its negation, written $\neg q(t_1, \ldots, t_n)$.
A rule has the form
\[
h \lpif b_1, \ldots, b_n.
\]
where $h$ is an atom and each $b_i$ is a literal. All variables are universally quantified. 
Informally, such a rule states that $h$ is true whenever all of the $b_i$ are true. 
Here, $\lpif$ denotes logical implication ($\leftarrow$), and the comma ($,$) denotes conjunction ($\wedge$). We will also allow a disjunction ($\vee$) using the semicolon symbol ($;$), which is semantically equivalent to having separate rules with the same head $h$.
Rules with no body ($n = 0$) are called \emph{facts}. 
A logic program is a finite set of rules. 
We adopt the Prolog naming convention: constants, functors, and predicates begin with a lowercase letter, while variables begin with an uppercase letter.
We assign a label~$l$ to an atom~$f$ using the notation:
\[
l\prob f 
\]


\bibliographystyle{apalike}
\bibliography{main.bib,MyLibrary,ourbib15}

\end{document}